\documentclass{ieeeblanco}

\usepackage{bm}
\usepackage{cite}
\usepackage{amsmath,amssymb,amsfonts}
\usepackage{algorithmic}
\usepackage{graphicx}
\usepackage{textcomp}

\usepackage{empheq}

\makeatletter
\AtBeginDocument{\DeclareMathVersion{bold}
\SetSymbolFont{operators}{bold}{T1}{times}{b}{n}
\SetSymbolFont{NewLetters}{bold}{T1}{times}{b}{it}
\SetMathAlphabet{\mathrm}{bold}{T1}{times}{b}{n}
\SetMathAlphabet{\mathit}{bold}{T1}{times}{b}{it}
\SetMathAlphabet{\mathbf}{bold}{T1}{times}{b}{n}
\SetMathAlphabet{\mathtt}{bold}{OT1}{pcr}{b}{n}
\SetSymbolFont{symbols}{bold}{OMS}{cmsy}{b}{n}
\renewcommand\boldmath{\@nomath\boldmath\mathversion{bold}}}
\makeatother

\def\BibTeX{{\rm B\kern-.05em{\sc i\kern-.025em b}\kern-.08em
    T\kern-.1667em\lower.7ex\hbox{E}\kern-.125emX}}

\usepackage{amssymb}

\usepackage{t1enc,amsmath,amsfonts,epsfig,graphicx,color,soul,mathabx,mathtools,listings}

\usepackage[table]{xcolor}
\usepackage{multirow}

\usepackage{hhline}
\newcommand{\bi}[1]{\bm{\textit{#1}}}
\newcommand{\eq}[1]{\begin{align}#1\end{align}}

\sethlcolor{red}

\input{def11.set}

\graphicspath{{./Figs/}{./fig/}}

\begin{document}
\history{Date of publication xxxx 00, 0000, date of current version xxxx 00, 0000.}
\doi{.../...}

\title{Generalized Exponentiated Gradient Algorithms
	and Their Application to On-Line Portfolio Selection}
\author{\uppercase{Andrzej Cichocki}\authorrefmark{1}, 
\uppercase{Sergio Cruces}\authorrefmark{2}, 
\uppercase{Auxiliadora Sarmiento}\authorrefmark{2},
\uppercase{Toshihisa Tanaka}\authorrefmark{3},
}

\address[1]{Systems Research Institute of Polish Academy of Science, Newelska 6, 01-447 Warszawa, Poland\\
Global Innovation Research Institute of Tokyo University of Agriculture and Technology, 2-24-16 Naka-cho, Koganei-shi, Tokyo, Japan\\ 
Riken AIP, 103-0027 Tokyo, Nihonbashi, 1 chome-4-1, Japan\\
(e-mail: cichockiand@gmail.com)}
\address[2]{Departamento de Teoría de la Señal y Comunicaciones, Universidad de Sevilla, Camino de los Descubrimientos s/n, 41092 Seville, Spain \\(e-mails: \{sergio,\ asarmiento\}@us.es)}
\address[3]{Department of Electrical Engineering and Computer Science, Tokyo University of Agriculture and Technology, 2-24-16 Naka-cho, Koganei-shi, Tokyo, Japan (e-mail: tanakat@cc.tuat.ac.jp)}
\tfootnote{
This work was supported by the grant PID2021-123090NB-I00, funded by MCIN/AEI/ 10.13039/501100011033 and by “ERDF A way of making Europe”, FEDER/Ministerio Ciencia, Innovación y Universidades, Spain. It was also supported by PAIDI Andalucian grants P20\_01173 and US-1264994/FEDER, EU. 
A.C. and T.T. have been supported by Institute Global Innovation Research  (GIC) Institute of Tokyo University of Agriculture and Technology (TAUT), Tokyo, Japan.
}

\markboth
{Preprint version}
{Preprint version}

\corresp{Corresponding author: Sergio Cruces (e-mail: sergio@us.es).}

\begin{abstract}
This paper introduces a novel family of generalized exponentiated gradient (EG) updates derived from an Alpha-Beta divergence regularization function. Collectively referred to as EGAB,
the proposed updates belong to the category of multiplicative gradient algorithms for positive data and demonstrate considerable flexibility by controlling iteration  behavior and performance  through three hyperparameters: $\alpha$, $\beta$, and the learning rate $\eta$. To enforce a unit $l_1$ norm constraint for nonnegative weight vectors within generalized EGAB algorithms, we develop two slightly distinct approaches. One method exploits scale-invariant loss functions, while the other relies on gradient projections onto the feasible domain. As an illustration of their applicability, we evaluate the proposed updates in addressing the online portfolio selection problem (OLPS) using gradient-based methods. Here, they not only offer a unified perspective on the search directions of various OLPS algorithms (including the standard exponentiated gradient and diverse mean-reversion strategies), but also facilitate smooth interpolation and extension of these updates due to the flexibility in hyperparameter selection. Simulation results confirm that the adaptability of these generalized gradient updates can effectively enhance the performance for some portfolios, particularly in scenarios involving transaction costs.
\end{abstract}


\titlepgskip=-21pt

\maketitle

\onecolumn
\section{Introduction}\label{sec:introduction}

\label{introduction}
\color{black} 
In the era of massive data, research on gradient-based optimization has attracted renewed interest, due to many practical applications.
Gradient updates currently play a key role in signal processing, machine learning, and artificial intelligence, especially in deep neural networks.

In the literature, several fundamental formulations of gradient updates have been considered: additive gradient descent (GD) \cite{GD} and its stochastic version (SGD), multiplicative updates (MU) \cite{Cia3} and mirror descent (MD) updates \cite{EGSD}. 
The exponentiated gradient (EG) descent update belongs to the class of multiplicative gradient updates \cite{Cia3} and simultaneously to mirror descent algorithms (see e.g., \cite{EGSD}--\!\cite{MD1}) and it has been first  introduced by Kivinen and Warmuth in \cite{EG} and \cite{KW1995} and  adopted  for various applications by many researchers \cite{Helmbold98}-\cite{EGquantum}.

The main objective of this paper is to derive a new generalized family of exponentiated gradient updates, which we denote as EGAB. They can be regarded as a natural extension and generalization of existing exponentiated gradient methods that offer improved performance and considerably higher flexibility to adapt to data with different distributions by suitably learning the best hyperparameters.
The proposal is first developed for unnormalized updates (referred to as $EGAB\mbox{-}U$) and subsequently extended to normalized updates through the algorithmic variants $EGAB\mbox{-}N$ and $EGAB\mbox{-}P$. 

In this paper, within the framework of learning or optimization of loss functions under regularization, we introduce the Alpha-Beta divergence regularizer (briefly AB-divergence), which will be able to generalize and unify multiplicative and additive gradient updates \cite{Cichocki_Cruces_Amari},\cite{Cich-Amari-entropy}. The scope of the approach presented in this paper is vast, since the AB-divergence function includes a large number of useful regularization functions  including, Kullback-Leibler divergence,  Euclidean distance, Hellinger distance, Jensen-Shannon divergence,  Pearson and Neyman Chi-square divergences and Itakura-Saito distance and many more.  Furthermore, it also provides a natural extension of the families of Alpha- and  robust Beta-divergences providing  smooth (continuous) connections between them and links to other fundamental divergences  \cite{Cichocki_Cruces_Amari}.

In contrast to the standard EG algorithms the proposed generalization of the exponentiated gradient descent provides a wide family of updates, which is able to smoothly interpolate between additive gradient descent and multiplicative gradient descent updates. 
In summary, the main contributions of this work are as follows:
\begin{itemize}
	\item We introduce a unified approach to regularize loss functions using the quite general Alpha-Beta~divergence.
	\item We derive generalized exponentiated gradient updates, whose hyperparameters allow a certain degree of flexibility in optimized performance and sparsity of the solution depending on the distribution of training data.	
	\item We discuss how the sparsity of weight vectors is affected by orthogonal and scaled projections, and the role played by the hyperparameters of the algorithms.
	\item The proposed formulations are quite versatile and can be used in many potential applications. Here, we show their efficacy in the problem of online portfolio selection (OLPS), where they cover and extend several of the existing gradient-based methods. 
	\item Within this application, we propose an improved criterion that accounts for the existence of transaction costs, and show the effectiveness of learning algorithmic hyperparameters and data preprocessing options through a validation set with a past sample fraction. Findings that have been corroborated through an extensive set of OLPS experiments.
\end{itemize}
The article is structured as follows. Section II provides a brief overview of the types of gradient descent updates central to our exposition. In Section III, we review the definition of the AB~divergence and its relevant cases. Section IV derives a generalized unnormalized $EGAB\mbox{-}U$ algorithm for the optimization of a differentiable loss function.  In Section V, we propose additional generalized $EGAB\mbox{-}N$ and $EGAB\mbox{-}P$ updates that are now designed to preserve the $\ell_1$~norm constraint of the solution. In Section VI, we illustrate the application of the proposed formulations in problem of on-line portfolio selection (OLPS). Section~VII includes simulation experiments that test the validity of the developed updates in comparison with other state-of-the-art gradient-based methods for OLPS. Finally, Section VIII presents the conclusion of this work. 

\section{Classical gradient descent algorithms}
\color{black}
The additive gradient descent (GD) update can be intuitively justified by minimizing suitably regularized loss function \cite{EG}
\eq{
	\bw_{t+1} = \argmin_{{\bi w} \in \mathbf{R}^N} \left\{ L(\bw )+\frac{1}{2\eta} \|\bw-\bw_t\|^2_2 \right\},
	\label{Eq-1}
}
where $\bw_t$ is a vector of weights with $\bw_t=[w_{1,t},\ldots,w_{N,t}]^T\in \mathbf{R}^N$.  $L(\bw)$ is a differentiable loss function and $\eta > 0$ is a learning rate.
Differentiating the above cost function and equating it zero we have so called implicit gradient descent:
\eq{
	\bw_{t+1}=\bw_t - \eta \nabla_{\bi w} L(\bw_{t+1}).
	\label{eq_initial}
}
The term implicit implies that update uses of the loss function $L(\bw)$ at a future point $\bw_{t+1}$ \cite{he2008explicit}. 
In practice, usually this is approximated by the explicit update 
\eq{
	\bw_{t+1}=\bw_t - \eta \nabla_{\bi w} L(\bw_{t}),
}
assuming that
$\nabla_{\bi w} L(\bw_{t+1}) \approx \nabla_{\bi w} L(\bw_{t})$. The approximation is valid under some general conditions \cite{GD}, but it is also apparent when one replaces the original loss $L(\bw)$ in \eqref{Eq-1} by its linear approximation at point $\bw_t$, i.e., 
\eq{
	\hat L(\bw) = L(\bw_t) + (\bw-\bw_t) \nabla_{\bi w} L(\bw_t).
}

For on-line portfolio selection, we can assume that weights are nonnegative  $w_i \geq 0$ for $i=1,2,\ldots,N$ and  additionally, typically scaled or  normalized to unit $\ell_1$-norm, i.e.,  $\|\bw\|_1 =\sum_{i=1}^N w_i=1$. 
In such case,  standard EG update can be derived by minimizing the following optimization  problem
\eq{
	\bw_{t+1} = \argmin_{{\bi w} \in \mathbf{R}_+^N} \left\{ \hat L(\bw)+\frac{1}{\eta}D_{KL} (\bw \| \bw_t) \right\},
}
where the Kullback--Leibler divergence for probability mass functions 
\eq{
	D_{KL} (\bw_{t+1} \| \bw_t) = \sum_{i=1}^N w_{i,t+1} \log \frac{w_{i,t+1}}{w_{i,t}} 
}
is employed as regularizer. By minimizing the unconstrained regularized cost function, we obtain the so called EG unnormalized update
\eq{
	\bw_\star =\bw_t \odot \exp [- \eta \nabla_{\bi w} L(\bw_{t})],
	\label{EG-part1}
}
where $\odot$ denotes component-wise (Hadamard) multiplication of two vectors.
The unnormalized solution $\bw_\star$ can be converted to a unit $\ell_1$-norm length vector through a proper scaling of the update after each iteration step
\eq{
	\bw_{t+1}= \bw_\star/\|\bw_\star\|_1.
	\label{EG-part2}
}
Therefore, the standard or normalized EG iteration is summarized by equations \eqref{EG-part1}-\eqref{EG-part2}. 
Alternatively, the EG algorithm can be also derived by optimizing the following Lagrangian function \cite{EG}
\eq{
	J(\bw_{t+1}) =  \hat{L}(\bw_{t})  + \frac{1}{\eta} D_{KL}(\bw_{t+1} \| \bw_{t}) + \lambda \left(\sum_{i=1}^{N} w_{i,t+1}-1\right),\nonumber
}
where $\lambda >0$ is the Lagrange multiplier. The saddle-point of this function leads to the standard EG algorithm, expressed now in scalar form as
\eq{
	w_{i,t+1} = w_{i,t} \; \frac{\exp [- \eta \nabla_{w_{i,t}} L(\bw_{t})]}{\sum_{j=1}^N w_{j,t} \exp [- \eta \nabla_{w_{j,t}} L(\bw_{t})]}, \quad  i=1,\ldots,N.
	\nonumber
}

It is noteworthy that the aforementioned categories gradient descent updates can be regarded as discrete approximations of continuous-time updates described by ordinary differential equations (ODE), specifically:
\begin{itemize}
	
	\item Gradient Descent (GD)
	\eq{
		\frac{d \bw(t)}{d t}= -\mu \nabla_{\bi w} L(\bw(t))
	}
	\item Unnormalized Exponentiated Gradient ($EGU$)
	\eq{
		\frac{d \ln\!\left(\bw(t)\right)}{d t}= -\mu \nabla_{\bi w} L(\bw(t))
		\label{ODE2}
	}
	\item Mirror Descent (MD)
	\eq{
		\frac{d\,f\!\left(\bw(t)\right)}{d t}= -\mu \nabla_{\bi w} L(\bw(t))
	}
\end{itemize}
where $\mu>0$ is a learning rate for continuous time learning and $f(\bw)$ is suitably chosen link function \cite{EGSD}--\!\cite{MD1}. In this sense, the EG update corresponds to the discrete-time version of continuous ODE in \eqref{ODE2}), obtained via Euler discretization 
\eq{
	\bw_{t+1} 
	&= \exp \left( \ln \bw_t -\mu \Delta t\, \nabla_{\bi w} L(\bw_t) \right) \nonumber \\
	&= \bw_t \odot \exp \left( - \eta \nabla_{\bi w} L(\bw_t) \right),\nonumber
}
where $\eta = \mu \Delta t >0$ is  a learning for discrete time updates, which must be sufficiently small to ensure convergence of the algorithm.
Note that EG updates belong to wider class mirror  descent (MD) algorithms, where the natural logarithm $f(\bw) = \ln(\bw)$ is chosen as link function \cite{MD2,MD3}. Indeed, numerous extensions and applications of unnormalized and normalized EG updates have been proposed in the current literature, see \cite{MD1}-\!\cite{Inverse} and \cite{EGMA}-\!\cite{Momentum} for an overview. 

\color{black}
\section{The Alpha-Beta family of divergences}\label{ABdiv1} 

In general, the incorporation of suitable divergences as regularization functions in the optimization criterion is an essential step to create algorithms with interesting properties. Alpha-Beta divergence is a family of smoothly connected divergences, which is parameterized by $(\alpha,\beta)\in \mathbf{R}^2$. These hyperparameters enable a smooth interpolation between several important families of well-known divergences, and can be also selectively chosen to provide different degrees of robustness with respect to noise and outliers in the observations, as it was already demonstrated in  \cite{Cichocki_Cruces_Amari}. To the best of our knowledge, the generalized EG updates that correspond to a two dimensional parameterization of the AB-divergence have never been proposed or studied so far. 
In order to make the article self-contained, we will provide a brief summary of the fundamental definitions of AB-divergence for positive measures. Readers interested in its properties and robustness aspects can refer to the original exposition in \cite{Cichocki_Cruces_Amari}.

Without loss of generality, we initially consider the case where the parameters $\alpha,\ \beta$, and $\alpha+\beta$, are nonzero, and later particularize these expressions to the special cases where $\alpha,\ \beta$, or $\alpha+\beta$, may be zero.
For two positive vectors $\bw_{t+1} \in \mathbf{R}_+^N $ and $\bw_t \in \mathbf{R}_+^N $ with unit $l_1$ norm, let us consider the following dissimilarity measure,
which we shall refer  to as the Alpha-Beta divergence or shortly AB-divergence:
\eq{
	D^{(\alpha,\beta)}_{AB}({\bw_{t+1}}\|{\bw_{t}})= \sum_{i=1}^N d^{(\alpha,\beta)}_{AB}(w_{i,t+1} \| w_{i,t}),
	\label{AB}
}
where each of these terms 
\eq{
	&d^{(\alpha,\beta)}_{AB}(w_{i,t+1}\| w_{i,t}) =\\
	&\hspace{0.5cm}
	- \displaystyle  \frac{1}{\alpha \beta} \displaystyle \left( w_{i,t+1}^\alpha w_{i,t}^\beta
	- \frac{\alpha}{(\alpha+\beta)} w_{i,t+1}^{\alpha+\beta}
	- \frac{\beta}{(\alpha+\beta)} w_{i,t}^{\alpha+\beta}
	\right) 
	\label{3Young}
	\nonumber
}
represents the scalar AB-divergence between the corresponding elements of the vectors at the $i^{th}$ position. Note that  Eq. \eqref{AB} is a divergence when $\alpha,\beta,\alpha+\beta\neq 0$, 
since it was shown in the Appendix of \cite{Cichocki_Cruces_Amari} that the following general inequality holds true $d^{(\alpha,\beta)}_{AB}(w_{i,t+1}\| w_{i,t}) \geq 0$, with equality holding only for $w_{i,t+1}=w_{i,t}$. 
\color{black}

\subsection{Extension by continuity of the AB-divergence}

In order to avoid indeterminacy or singularity for certain values of parameters where $\alpha,\beta$ or $\alpha+\beta$ can be zero, the AB-divergence extends its definition by continuity (by applying l'H\^opital's rule) being then well defined across the $(\alpha,\beta)\in \mathbf{R}^2$ plane. 

For $\alpha\neq 0,\ \beta = 0 $, one obtains the generalized family of Kullback-Leibler divergences 
\eq{\small
	D^{(\alpha,0)}_{AB}(\bw_{i,t+1}\| \bw_{i,t})=
	\frac{1}{\alpha^2} \sum_{i=1}^N \displaystyle 
	\displaystyle \left( w_{i,t+1}^\alpha \ln  \displaystyle \frac{w_{i,t+1}^\alpha}{w_{i,t}^\alpha} - w_{i,t+1}^{\alpha} + w_{i,t}^{\alpha}  \right),
	\nonumber
}
which is parameterized by $\alpha\in \mathbf{R}\setminus \{0\}$. Simplifying into the Kullback-Leibler divergence \cite{Kullback} in the special case of $\alpha=1$.

Similarly, for $\beta\neq 0,\alpha = 0 $, one obtains the generalized dual family of Kullback-Leibler divergences 
\eq{
	D^{(0,\beta)}_{AB}(\bw_{i,t+1}\| \bw_{i,t})= 
	\frac{1}{\beta^2} \sum_{i=1}^N \displaystyle  
	\left( w_{i,t}^\beta \ln  \displaystyle \frac{w_{i,t}^\beta}{w_{i,t+1}^\beta} - w_{i,t}^{\beta} + w_{i,t+1}^{\beta}  \right)
	\nonumber
}
which is parameterized by $\beta\in \mathbf{R}\setminus \{0\}$. The classical dual Kullback-Leibler divergence is obtained for $\beta=1$.

For $\alpha=-\beta> 0$, one obtains a generalized family of Itakura-Saito divergences, 
\eq{
	D^{(\alpha,-\alpha)}_{AB}({\bw_{t+1}}\| \,{\bw_{t}})
	=
	\frac{1}{\alpha^2} \displaystyle \sum_{i=1}^N
	\left(\ln \frac{w_{i,t}^\alpha}{w_{i,t+1}^\alpha} + \frac{w_{i,t+1}^\alpha}{w_{i,t}^\alpha}  -1 \right) 
	\nonumber
}
which is parameterized by $\alpha\in \mathbf{R}\setminus \{0\}$, and simplifies for $\alpha=1$ and $\beta=-1$ into the standard Itakura-Saito divergence $D_{IS}({\bw_{t+1}}\| \,{\bw_{t}})$ (see \cite{IS}).

The last indeterminacy happens when $\alpha=\beta=0$, case where the AB-divergence simplifies into the popular Log-Euclidean metric \cite{logEuclidean} 
\eq{
	D^{(0,0)}_{AB}({\bw_{t+1}}\| \,{\bw_{t}})
	=
	\frac{1}{2} \displaystyle\sum_{i=1}^N
	\left(\ln w_{i,t+1} - \ln w_{i,t} \right)^2.
}

\subsection{Special important cases for AB-divergence}

The designation of the Alpha-Beta divergence is explained by the fact that, through a common expression, it is able to unify and continuously interpolate two popular families of divergences: the Alpha divergences \cite{Cich-Amari-entropy}--\!\cite{Minka05} 
and the Beta divergences \cite{Basu}--\!\cite{CichZd_ICAISC06}. For $\beta=1-\alpha$, the AB-divergence is particularized to the well-known family of Alpha divergences 
\eq{
	&D^{(\alpha,1-\alpha)}_{AB} \left({\bw_{t+1}} \|  {\bw_{t}}\right)
	=
	D^{(\alpha)}_{A}\left({\bw_{t+1}}\|{\bw_{t}}\right)  \label{DA1} \\ 
	&\hspace{.2cm} \doteq
	\left\{
	\begin{tabular}{l}
		$\displaystyle \frac{1}{\alpha(\alpha-1)} \displaystyle \sum_{i} \left( w_{i,t+1}^\alpha w_{i,t}^{1-\alpha}
		- \alpha w_{i,t+1} +(\alpha-1) w_{i,t} \right)  $  \\ \hspace{4.5cm} $\ \text{for}\ \alpha\in  \mathbf{R}\setminus \{0,1\} $
		\\
		$\displaystyle \sum_{i} \left( w_{i,t+1} \ln \displaystyle \frac{w_{i,t+1}}{w_{i,t}}
		- w_{i,t+1} + w_{i,t}
		\right) $ \ \  $\ \text{for}\ \alpha=1 $
		\\
		$\displaystyle \sum_{i} \left( w_{i,t} \ln \displaystyle \frac{w_{i,t}}{w_{i,t+1}}
		- w_{i,t} + w_{i,t+1}
		\right) $  \ \ \ \ \ \ $\ \text{for}\ \alpha= 0 $.
	\end{tabular}
	\right.
	\nonumber
}

Whereas, for $\alpha=1$, the AB-divergence it reduces to the Beta-divergence 
\eq{
	&D^{(1,\beta)}_{AB} \left({\bw_{t+1}}   \|  {\bw_{t}}\right)  =  D^{(\beta)}_{B}({\bw_{t+1}}\|{\bw_{t}}) \label{betadiv1}\\
	&\hspace{.1cm} \doteq
	\left\{
	\begin{tabular}{l}
		$- \displaystyle \frac{1}{\beta}\sum_{i} \left( w_{i,t+1} w_{i,t}^\beta
		- \frac{w_{i,t+1}^{1+\beta}}{(1+\beta)}
		- \displaystyle \frac{\beta w_{i,t}^{1+\beta}}{(1+\beta)} 
		\right)  $ \\
		\hspace{4.5cm}$\ \text{for}\ \beta\in  \mathbf{R}\setminus \{-1,0\} $
		\\
		$\displaystyle \frac{1}{2} \sum_{i} \left( w_{i,t+1} -w_{i,t}
		\right)^2 $ \hspace{2.5cm}  $\ \text{for}\ \beta= 1 $
		\\
		$\displaystyle \sum_{i} \left( w_{i,t+1} \ln \displaystyle \frac{w_{i,t+1}}{w_{i,t}}
		- w_{i,t+1}  + w_{i,t}
		\right) $  \hspace{.1cm}   $\ \text{for}\ \beta= 0 $
		\\
		$\displaystyle \sum_{i} \left( \ln \displaystyle \frac{w_{i,t}}{w_{i,t+1}}+
		\left(\displaystyle \frac{w_{i,t}}{w_{i,t+1}}\right)^{-1}-1
		\right)  $ \hspace{.3cm}   $\ \text{for}\ \beta=-1 $.
	\end{tabular}
	\right.
	\nonumber
}

Particular cases of the AB-divergence when $\alpha=\beta$ include: the squared Euclidean distance (for $\beta=1$), the Hellinger distance  (for $\beta=0.5$), and the log-Euclidean distance (for $\beta=0$).
In general, when $\alpha=\beta$, the AB-divergence is symmetric with respect to both arguments and takes the form of a generalized Euclidean metric distance 
\eq{ \label{metric}
	D^{(\beta,\beta)}_{AB}({\bw_{t+1}}\|{\bw_{t}})
	&=
	D_{E}(\log_{1-\beta}({\bw_{t+1}})\|\log_{1-\beta}({\bw_{t}}))\\
	&\dot=
	\frac{1}{2}\sum_{i=1}^N \left( \log_{1-\beta}(w_{i,t+1}) - \log_{1-\beta}(w_{i,t}) \right)^2 
}
evaluated in the image domain of a link function that equals the $1-\beta$ deformed logarithm of Tsallis \cite{Tsallis}. For $x>0$, this is defined as the following Box-Cox power transformation \cite{BoxCox-power} with parameter $\beta$ 
\eq{ \label{deflog}
	\log_{1-\beta}(x)=\left\{
	\begin{array}{cl}
		\displaystyle \frac{x^{\,\beta}-1}{\beta}  & \beta\neq 0,\\
		\ln(x) & \beta=0.
	\end{array}
	\right.
}

The inverse of the link function is the deformed or generalized exponential $\exp_{1-\beta} (x)$, which is defined as follows
\eq{ \label{defexp}
	\exp_{1-\beta}(x)=\left\{
	\begin{array}{cl}
		\displaystyle [1+ \beta x]_+^{1/\beta} & \beta\neq 0\\
		\exp(x) & \beta=0
	\end{array}
	\right.
}
\begin{figure}[]
	\begin{center}
		\includegraphics[width=.7\linewidth]{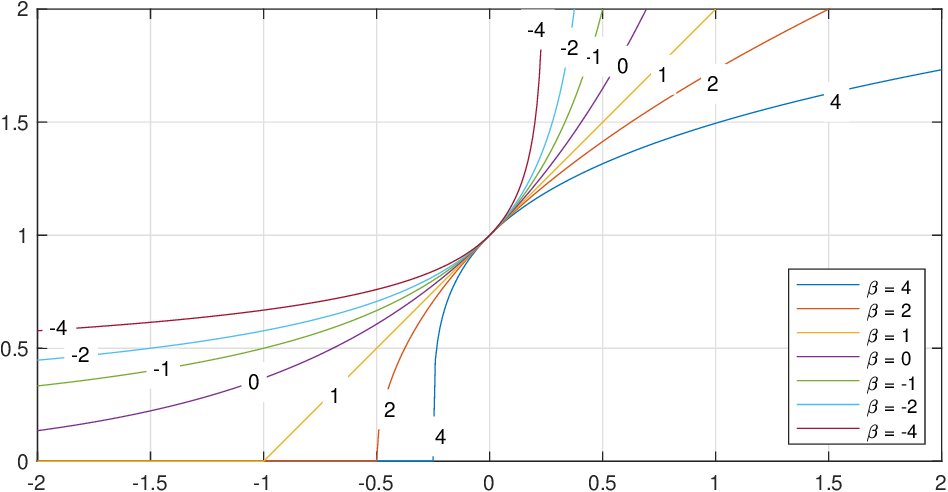}
		\caption{Plots of the generalized exponential function $\exp_{1-\beta}(x)$ for different values of parameter $\beta$.}
		\label{Fig-def-exp}
	\end{center}
\end{figure}
Both the deformed logarithms and exponentials play a key role in derivation of the proposed generalized EG updates.

\section{Unnormalized exponentiated gradient updates for AB-divergence regularizers} \label{ABEG1}

In this section, we adapt the approach developed in derivation of standard EG updates to present the generalized and unnormalized EG iteration based on the wide class of Alpha-Beta divergences. 
Our objective is to find  a new vector $\bw_{t+1} =[w_{1,t+1},w_{2,t+1},\ldots, w_{N,t+1}]^T \in \mathbf{R}_+^N $ that solves the following optimization problem
\eq{
	\bw_{t+1} = \argmin_{{\bi w} \in \mathbf{R}_+^N} \left \{
	L(\bw)+\frac{1}{\eta} D^{(\alpha,\beta)}_{AB}(\bw\|\bw_{t})
	\right\}\; .
	\label{problem3a}
}
It should be emphasized that the AB-divergence serves as a flexible regularizer  or a penalty term which tends to keep $\bw_{t+1}$ as close as possible to $\bw_t$. The adaptive learning rate $\eta_t>0$ controls  the  balance or relative importance between the loss function $L(\bw) $ and the regularizer  $D^{(\alpha,\beta)}_{AB}(\bw \| \bw_t)$.

For obtaining an approximate local solution in close-form, we should reinterpret  $L(\bw)$ as a composition of functions 
\eq{
	L(\bw)=L((\bw^\alpha)^{1/\alpha})=F(\bw^\alpha),
}
which inner argument is $\bw^\alpha\equiv[w_1^\alpha,\ldots,w_N^\alpha]^T$, where the power of the vectors are expected to act component-wise on their elements. Then one can locally approximate the value of $F(\bw^\alpha)$ by first-order Taylor expansion $\widehat{F}(\bw^\alpha)$ in the neighborhood of $\bw_t^\alpha$, i.e.,
\eq{
	\widehat{F}(\bw^\alpha) = F(\bw_{t}^\alpha) +  \langle\ \nabla_{{\bi w}^\alpha} F(\bw_t^\alpha),\ (\bw^\alpha-\bw_t^\alpha)\ \rangle .
	\label{Fdef}
}
\color{black}
For the sake of clarity, we will omit our derivations of the generalized EGU update in the main text, and refer the interested reader to Appendix \ref{App-EG-AB}. There, we explain in detail how the solution of the approximate optimization problem
\eq{
	\bw_{t+1} = \argmin_{{\bi w} \in \mathbf{R}_+^N} \left \{
	\widehat{F}(\bw^\alpha)
	+ \frac{1}{\eta} D^{(\alpha,\beta)}_{AB}(\bw\|\bw_{t})
	\right\}
	\label{problem3b}
}
leads to the proposal of the general unnormalized update   
\eq{
	w_{i,t+1} = w_{i,t} \exp_{1-\beta} \left(- \eta_{i,t}  \frac{\partial L(\bw_{t})}{\partial w_i} \right),
	\label{BG-EGU}
}
for $i=1,\ldots,N$, with the time-varying learning rate
\eq{
	\eta_{i,t} = \eta\, w_{i,t}^\gamma > 0.
	\label{step1}
}
Note that all dependence of the step-size on $(\alpha,\beta)$ is summarized into the key parameter
\begin{empheq}[box=\fbox]{align}	
	\gamma=1-(\alpha+\beta)\, .
\end{empheq}  
The unnormalized $EGAB\mbox{-}U(\alpha,\beta)$  update in vector form is given by
\begin{empheq}[box=\fbox]{align}	
	\bw_{t+1}
	= \bw_{t} \odot \exp_{1-\beta} \left(-\bm{\eta}_t \odot \nabla_{\bi w}L(\bw_t) \right)
	\label{sol1b}
\end{empheq} 
where $\bm{\eta}_t=\eta\, \bw_t^\gamma$ is a vector of learning rates, which components have been previously defined in \eqref{step1}. 

\subsection{On the role of hyperparameters controlling the iteration process}

The three parameters $\beta$, $\gamma$ and $\eta$, govern different aspects of the behaviour of the $EGAB\mbox{-}U(\alpha,\beta)$ iterations. As we will see, while $\beta$ controls the mass-covering and mode-seeking behaviour of the weight distribution, both $\gamma$ and $\eta$ control different aspects of the vector of learning rates $\bm{\eta}_t$.  $\gamma$ regulates the influence of the current estimates in the learning rates, and $\eta$ may be regarded as a common step-size that controls the magnitude of  $\bm{\eta}_t$. This is explained in more detail below.

Since the weights $w_{i,t}$ are positive, 
we can rewrite the estimating equation \eqref{BG-EGU} as
\eq{
	\log_{1-\beta}\left(\frac{w_{i,t+1}}{w_{i,t}}\right)= - \eta\, 	w_{i,t}^{\gamma}\,  \frac{\partial L(\bw_{t})}{\partial w_i} .
	\label{zoom}
}
As it can be observed in the left-hand-size of \eqref{zoom}, the estimation equation resulting from the optimization of AB-divergence regularizers involves here a $\beta$-zoom (see \cite{Cichocki_Cruces_Amari}) of the ratios $w_{i,t+1}/w_{i,t}$ through the function $\log_{1-\beta}(\cdot)$. 

The parameter $\beta\in \mathbf{R}$ determines the degree of convexity of the deformed logarithm, which is concave for $\beta<1$, linear for $\beta=1$, and convex for $\beta>1$. Therefore, $\beta$ can be used to control the relative importance of large ratios over small ones in the estimating equations, and vice versa. For $\beta<1$, the relative relevance of smaller ratios (the best fits where $w_{i,t+1}\approx w_{i,t}$ or when $w_{i,t+1}\ll w_{i,t}$) is emphasized over the larger ones (the bad fits where $w_{i,t+1}\gg w_{i,t}$). This promotes a mass covering-behaviour, with a tendency for the new weights $\bw_{t+1}$ to approximately fit or largely surpass in value the older ones in $\bw_t$. This often favors a solution where a few weights can dominate the portfolio. On the contrary, for $\beta>1$ the higher ratios are relatively more relevant, which favors a mode-seeking behavior, the tendency for the weight of $\bw_{t+1}$ to match or fall below those of $\bw_t$, which usually results in a more uniform solution. 
Both the mass-covering and mode-seeking behavior are well-established concepts described by Minka \cite{Minka05} in the context of $\alpha$-divergences. 

One can observe in \eqref{zoom} how $\gamma\in \mathbf{R}$ controls the weighting of the gradient by $w_{i,t}^{\gamma}$. This parameter is an exponent that determines the influence of the current solution $\bw_t$ on the vector of learning rates $\bm{\eta}_t=\eta\, \bw_t^\gamma$. On the one hand, positive values of $\gamma$ exponentially slow down the learning rates of the smallest elements of $\bw_{t}$, which provides some degree of robustness against noisy updates on the small weights. On the other hand, negative values of $\gamma$ slow down the learning rate of the largest elements, which can be seen as a protection against very bad fits or outliers. Finally, for $\gamma=0$, the learning-rates are independent of $\bw_t$. In general, a suitable  minimum positive threshold value on the elements of $\bw_t$ is used to prevent both the existence numerical errors in the evaluation of $\bw_{t}^\gamma$ when $\gamma<0$, and the stopping of the updating of the null coefficients when $\gamma>0$.
For further information on the robustness to noise and outliers of AB~divergence based criteria we refer the interested reader to \cite{Cichocki_Cruces_Amari,Regli-Silva,GAN-AB}.

\section{Normalized exponentiated gradient updates based on AB-divergence regularizers} \label{NABEG1}
In the previous section, we derived the generalized $EGAB\mbox{-}U$ update. In this section, we propose the normalized variants of the exponentiated gradient updates $EGAB\mbox{-}N$ and $EGAB\mbox{-}P$, together with the justification and motivation for just such normalizations. These updates, which can be used to preserve the $\ell_1$ norm constraint of $\bw_{t+1}$, are respectively based on a gradient descent iteration for scale-invariant functions, and on a projected gradient descent iteration.  

\subsection{Normalization by scaling: the $EGAB\mbox{-}N$ update}

This generalized exponentiated gradient update is mainly suitable for the optimization of scale-invariant 
loss functions $L_I(\bw)$, i.e., functions that satisfies the following property
\eq{
	L_I(c\, \bw)= 
	L_I(\bw)\quad \forall c>0.
}
Even when this is not satisfied for a given differentiable loss function $L(\bw)$, we can obtain its scale-invariant version $L_I(\bw)$ through the simple normalization of the function arguments 
\eq{
	L_I(\bw)=L\left(\frac{\bw}{\|\bw\|_1}\right).
	\label{LI}
}

The proposed two-step normalized $EGAB\mbox{-}N(\alpha,\beta)$ update is   
\begin{empheq}[box=\fbox]{align}  
	\bw_{\star} &= \bw_{t} \odot \exp_{1-\beta}\left(-\bm{\eta}_t\odot 
	\nabla_{\bi w} L_I(\bw_{t})\right) \label{Iter1-7}\\
	\bw_{t+1}   &=  \bw_{\star}/\|\bw_{\star}\|_1 \, ,\label{Iter2-7}
\end{empheq}	
the sequential combination of the unnormalized update $EGAB\mbox{-}U(\alpha,\beta)$   in \eqref{Iter1-7} with the $\ell_1$-normalization of \eqref{Iter2-7}. In this update,  $\bm{\eta}_t=\eta\, \bw_t^\gamma$ is the vector of learning rates, and the required gradient can be evaluated in terms of  $\nabla_{\bi w} L(\bw_{t})$, since
\eq{
	\nabla_{\bi w} L_I(\bw_{t})
	&=\nabla_{\bi w} L(\bw_t) - (\bw_t^T\nabla_{\bi w} L(\bw_t))\, {\bf 1}  ,
	\label{eq_gradLI}
}
where ${\bf 1}=(1,\ldots,1)^T$ denotes a vector of unit elements.

We refer the readers interested in a detailed derivation of the update to the explanation presented in Appendix \ref{App-EGn}. There, with help of Euler's homogeneous function theorem, we show that the gradient $\nabla_{\bi w} L_I(\bw_t)$ is always orthogonal to the current iterate $\bw_t$ and, because of the scale invariance of the loss function, the $\ell_1$ normalization of $\bw_\star$ obtained in \eqref{Iter1-7} does not modify the value $L_I(\bw_t)$. 

Also, in Appendix \ref{App-2step}, we prove that the iteration can be equivalently summarized by the following multiplicative expression
\eq{\small
	\!\!\!\!\bw_{t+1}\!= \!  \bw_{t} \odot \exp_{1-\beta} \left(\frac{\bm{-\eta}_t\odot \nabla_{\bi w}L_I(\bw_t)}{\|\bw_{\star}\|_1^{\beta}}  \!+\!\log_{1-\beta}\|\bw_{\star}\|_1^{-1}\right)
	\label{single_update}
}
where $
\|\bw_{\star}\|_1 = \bw_{t}^T \exp_{1-\beta} \left(-\bm{\eta}_t \odot \nabla_{\bi w}L_I(\bw_t) \right)$. 

We next show how the proposal $EGAB\mbox{-}N(\alpha,\beta)$ can be particularized to two well-known gradient updates: $EGAB\mbox{-}N(1,0)$ which implements the classical multiplicative Exponentiated Gradient descent, and $EGAB\mbox{-}N(1,1)$ which provides the  additive normalized Gradient Descent. 

\subsubsection{Particularization to the Exponentiated Gradient descent}
For $\alpha=1$ and $\beta=0$, the Alpha-Beta divergence regularizer reduces to the Kullback-Leibler divergence and the proximal optimization of the loss function $L_I(\bw)$ can be performed by the Exponentiated Gradient algorithm. In this case, we have $\gamma=1-\alpha-\beta=0$, and the time-varying learning rate equalizes for all its modes, being equal to
\eq{
	\bm{\eta}_{t} = \eta\, \bm{1}.
}
The resulting $EGAB\mbox{-}N(1,0)$ update in \eqref{Iter1-7}-\eqref{Iter2-7} particularizes to the standard normalized version of the $EG$ algorithm (see \cite{Helmbold98}--\!\cite{EGnoise})
\eq{
	\bw_{\star} &= \bw_{t} \odot \exp \left(- \eta \nabla_{\bi w}L(\bw_t) \right)\label{Iter6-1a}\\
	\bw_{t+1}   &=  \bw_{\star}/\|\bw_{\star}\|_1 \,. \label{Iter6-2a}
}
Note that in \eqref{Iter6-1a}, the gradient $\nabla_{\bi w}L_I(\bw_t)$ within the deformed exponential has been replaced by $\nabla_{\bi w}L(\bw_t)$. This was possible because the offset between them, see \eqref{eq_gradLI}, is here just a constant term. This constant in the exponent simply scales $\bw_{\star}$ by a factor which later cancels out with the normalization in \eqref{Iter6-2a}.
According to the simplification of \eqref{single_update} for this case, the update can also be condensed into the expression
\eq{
	\bw_{t+1} &= \bw_{t} \odot \exp \left(- \eta \nabla_{\bi w}L(\bw_t)  
	+\log \|\bw_{\star}\|_1^{-1}\right)
}

\subsubsection{Particularization to the $\ell_1$-normalized Gradient Descent}

For $\alpha=1$ and $\beta=1$, the AB-divergence regularizer coincides with the Euclidean divergence. 
In this special case, $\gamma=-1$, and the  $EGAB\mbox{-}N(1,1)$ update in \eqref{Iter1-7}-\eqref{Iter2-7} simplifies to the $\ell_1$-normalized version ($GD\mbox{-}N$) of the Gradient Descent algorithm
\eq{
	\bw_{\star} 
	&= \bw_{t} -  \eta \nabla_{\bi w}L_I(\bw_t)  \label{Iter6-1g.}\\
	\bw_{t+1}   &=  \bw_{\star}/\|\bw_{\star}\|_1 \, \label{Iter6-2g}
}
for the optimization of $L_I(\bw)$. This is mainly due to the fact that for $1-\beta=0$ the deformed exponential is a linear function of its argument
\eq{
	\exp_{0} \left(-\bm{\eta}_t \odot \nabla_{\bi w}L_I(\bw_t) \right)
	= \bm{1}+\bm{\eta}_t \odot \nabla_{\bi w}L_I(\bw_t)\, ,
}
with $\bm{\eta}_{t} = \eta\, \bw_{t}^{-1}$.
This simplifies the multiplicative expression \eqref{Iter1-7} into the additive update \eqref{Iter6-1g.}. According to \eqref{single_update} the update in \eqref{Iter6-1g.}-\eqref{Iter6-2g} can be also expressed as
\eq{\small
	\bw_{t+1}\!= \!  \bw_{t} \odot \exp_{0} \left(-\frac{\eta\, \bw_{t}^{-1}}{\|\bw_{\star}\|_1} \odot \nabla_{\bi w}L_I(\bw_t) +\log_{0}\|\bw_{\star}\|_1^{-1}\right).
	\label{single_update2}
}

\subsubsection{Interpolated updates and other generalizations}

The $EGAB\mbox{-}N(1,\beta)$ update for $\beta\in[0,1]$, may be seen as a continuous interpolation between the $EGAB\mbox{-}N(1,0)$ algorithm (for $\beta=0$) and the $GD\mbox{-}N$ algorithm (for $\beta=1$). While different extrapolations and novel updates are obtained when other hyperparameters are substituted in the $EGAB\mbox{-}N(\alpha,\beta)$ proposal.

\subsection{Normalization by projection: $EGAB\mbox{-}P$ update} \label{ABEG3}

\begin{figure}[t]
	\begin{center}
		\includegraphics[width=.6\linewidth]{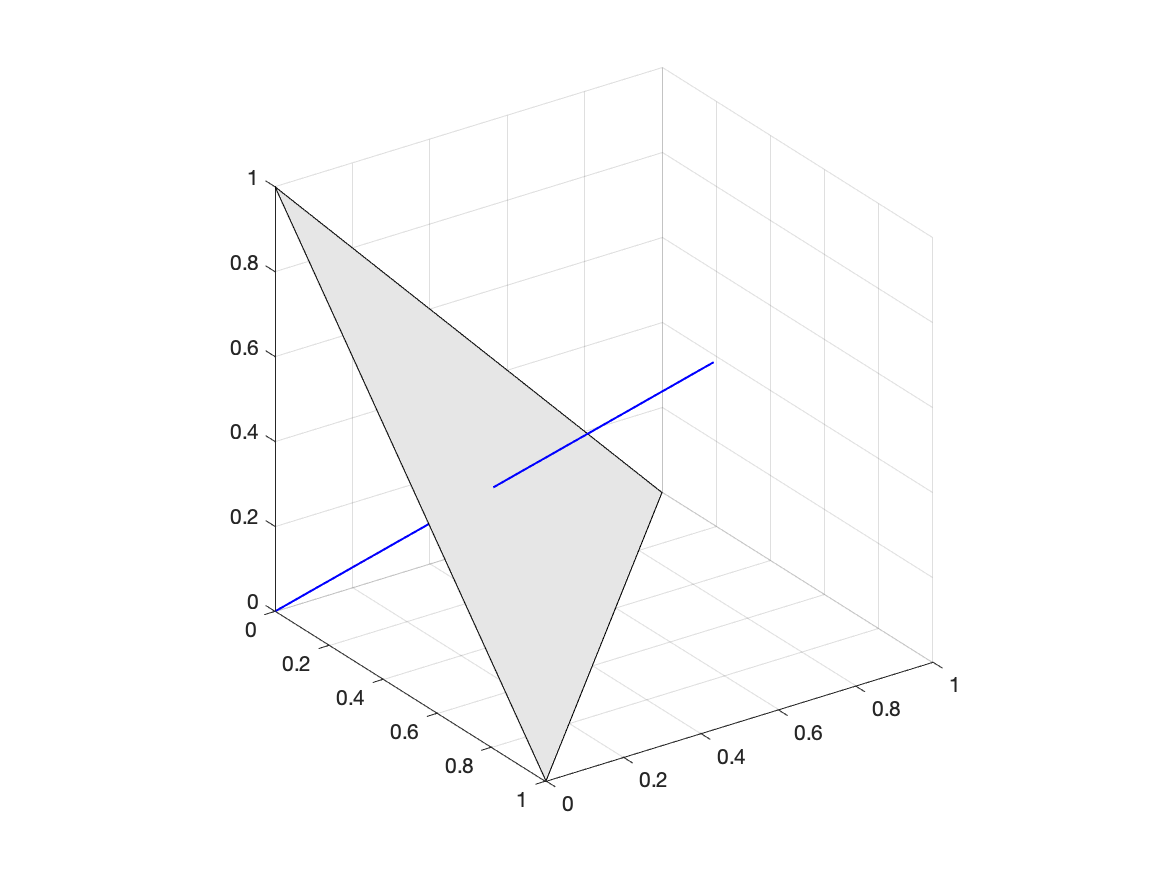}
		\caption{Visualization, in three  dimensions, of the $l_1$-norm constraint for nonnegative vectors. In blue color it is shown the orthogonal vector to the manifold which intersects with it at the uniform portfolio $\bu \dot=\tfrac{1}{N}{\bf 1}$.}
		\label{Fig0-simplex}
	\end{center}
\end{figure}

In this section, we introduce an alternative postprocessing of the unnormalized $EGAB\mbox{-}U$ update, aiming to ensure the unit value of  $\|\bw_{t+1}\|_1$. This will be attained at the feasible manifold shown in Figure~\ref{Fig0-simplex}. 
Let's start realizing that, for small changes in $\bw_{t+1}$, the increment of loss $L_I(\cdot)$ is dominated by its linear contribution
\eq{
	\Delta L_I(\bw_{t+1}) = \langle \nabla_{\bi w} L_I(\bw_t) ,\ \bw_{t+1}-\bw_t \rangle ,
	\label{delta1}
}
where the gradient $\nabla_{\bi w} L_I(\bw_t)$ admits a natural decomposition 
\eq{
	\nabla_{\bi w} L_I(\bw_t) = \nabla_{\bi w} L_\parallel(\bw_t) +  \nabla_{\bi w} L_{\perp}(\bw_t),
	\label{nabla1}
}
into parallel and orthogonal components to the feasible set of non-negative vectors with constant $l_1$-norm.

As we explain in Appendix \ref{App_decomp}, for nonnegative vectors $\bw_{t+1}$ and $\bw_t$ of unit $l_1$-norm, their difference  $\Delta \bw_{t+1}=\bw_{t+1}-\bw_t$ is orthogonal to $\nabla_{\bi w} L_{\perp}(\bw_t)$. Thus, with the substitution of \eqref{nabla1} into \eqref{delta1}, we realize that for small updates
\eq{
	L_I(\bw_{t+1})
	\approx L_I(\bw_{t}) + \langle \nabla_{\bi w} L_\parallel(\bw_t) ,\Delta \bw_{t+1} \rangle  
	\ \dot=\  L_\parallel(\bw_{t+1}) \label{Lparallel}
}
and the effective linear contribution of the increment in the loss function turns out to be dominated by the parallel component
\eq{
	\Delta L_I(\bw_{t+1})
	=\ \langle \nabla_{\bi w} L_\parallel(\bw_t) ,\Delta \bw_{t+1} \rangle \ =\ \Delta L_\parallel(\bw_{t+1})\  , \nonumber
}
which, is shown in Appendix \ref{App_decomp} to be given by
\eq{
	\nabla_{\bi w} L_\parallel(\bw_t) 
	&=  \nabla_{\bi w} L(\bw_t)- (\bu^T \nabla_{\bi w} L(\bw_t)) {\bf 1}.
	\label{Grad-Lpar}
}
Note also that, in \eqref{Grad-Lpar}, the convex combination of the gradient elements through the uniform vector $\bu \dot=\tfrac{1}{N}{\bf 1}$ coincides with the average gradient value of the initial loss function, i.e.,
\eq{
	\bu^T \nabla_{\bi w} L(\bw_t)  \equiv \overline{\nabla_{\bi w} L(\bw_t)}.
}
Hence, the following projected gradient update $EGAB\mbox{-}P(\alpha,\beta)$
\begin{empheq}[box=\fbox]{align}
	\bw_{\star} &= \bw_{t} \odot \exp_{1-\beta} \left(-\bm{\eta}_t\odot  \nabla_{\bi w} L_\|(\bw_{t}) \right) \label{AB-EG-proj_iter-1}\\
	\bw_{t+1} &= \text{\small projection onto the unit $l_1$-norm simplex}(\bw_{\star}), \label{AB-EG-proj_iter-2}
\end{empheq} 
with $\bm{\eta}_t=\eta\, \bw_t^\gamma$, is proposed to optimize the effective component $\Delta L_\parallel(\bw)$ of the loss function $L_I(\bw)$. 

The two-step update in \eqref{AB-EG-proj_iter-1} combines the  $EGAB\mbox{-}U$ iteration for the optimization of $\Delta L_\parallel(\bw)$, with the projection in  \eqref{AB-EG-proj_iter-2} of the intermediate solution $\bw_{\star}$ onto the unit $l_1$-norm simplex. This projection is chosen to promote the sparsity of the solution, being and scaled normalization when $\|\bw_{\star}\|\leq 1$ and the minimum distance projection otherwise. The details of its implementation are provided in Listings 1 and 2. 

\lstdefinestyle{mystyle}{
	backgroundcolor=\color{white},   
	commentstyle=\color{black},
	keywordstyle=\color{magenta},
	numberstyle=\tiny\color{black},
	stringstyle=\color{codepurple},
	basicstyle=\ttfamily\footnotesize,
	breakatwhitespace=false,         
	breaklines=true,                 
	captionpos=b,                    
	keepspaces=true,                 
	numbers=left,                    
	numbersep=5pt,                  
	showspaces=false,                
	showstringspaces=false,
	showtabs=false,                  
	tabsize=2
}
\begin{figure}[t]
	\color{black}
	\lstset{style=mystyle}
	\begin{lstlisting}[language=Matlab, caption=Projection of $\bw$ onto the unit $\ell_1$-norm simplex implemented in MatLab. The code of the minimum  distance projection is given in Listing~2.]
		function w_projected = projection(w)
		% Returns a sparse promoting projection of vector w onto the unit 1-norm simplex of 1st orthant. When norm(w,1)>1 the min. distance projection tends to be sparser than a normalization by scaling of w. On the contrary, for  norm(w,1)<1, the normalization by scaling tends to be sparser.   
		if norm(w,1)>1 
			w_projected = min_dist_projection(w);
		else
			w_projected = w/norm(w,1);
		end	
	\end{lstlisting}
	\lstset{style=mystyle}
	\begin{lstlisting}[language=Matlab, caption=Minimum distance projection of $\bw$ onto the unit $\ell_1$-norm simplex.]
		function w_projected = min_dist_projection(w)
		% Returns the projection of the vector w onto the unit 1-norm simplex of the 1st orthant. 
		start=1;
		while any(w<0) || start==1;
			start  =0;
			w(w<0) =0;   % Truncates negative elements.
			idx=find(w>0); % Ind. of positive elements.
			N  =length(idx); % N. of positive elements.
			one=ones(N,1)/sqrt(N); %Unit vector of 1s.   
			w(idx) = w(idx) +one/sqrt(N) ...
					     -(transpose(w(idx))*one)*one; 
			% Projection onto the plane orthogonal to the vector one, passing through the point one/sqrt(N). Since some of the projected coordinates may be negative, the previous procedure is iterated through the while loop until all the elements are nonnegative.   
		end
	\end{lstlisting}
\end{figure}

\subsection{On the behaviour of the generalized updates} 

The updates proposed in the previous sections, have a striking similarity and also certain differences that may leave a slight imprint on their behaviour. 

While $EGAB\mbox{-}N$ is a suitable technique for the optimization of scale-invariant functions, $EGAB\mbox{-}P$ was justified on the basis of a projected gradient method.   
Their multiplicative updates in \eqref{Iter1-7} and \eqref{AB-EG-proj_iter-1} share in common the functional form of $EGAB\mbox{-}U$ update
\eq{
	\bw_{\star} &= \bw_{t} \odot \exp_{1-\beta} \left(-\eta\, \bw_t^\gamma\odot   \nabla_{\bi w} L_\triangle(\bw_{t}) \right) , \label{AB-EG-proj_iter-1d}
}
with $\nabla_{\bi w} L_\triangle(\bw_{t})$ being equal to $\nabla_{\bi w} L_I(\bw_{t})$ in \eqref{Iter1-7}, and to $\nabla_{\bi w} L_\parallel(\bw_t)$ in \eqref{AB-EG-proj_iter-1}. Gradients that decompose similarly as
\eq{
	\left\{
	\begin{array}{l}
		\displaystyle
	\nabla_{\bi w} L_I(\bw_{t}) =   \nabla_{\bi w} L(\bw_t)- (\bw_t^T \nabla_{\bi w} L(\bw_t))\, {\bf 1} \, 
	\label{eq_gradLIc}\\
	\nabla_{\bi w} L_\parallel(\bw_t) =  \nabla_{\bi w} L(\bw_t)- (\bu^T \nabla_{\bi w} L(\bw_t))\, {\bf 1}
	\, .
	\end{array}
	\right.
}
Both expressions evaluate the excess of $\nabla_{\bi w} L(\bw_t)$ with respect to some constant thresholds resulting from convex combination of the gradient elements.  $EGAB\mbox{-}N$ uses the current vector $\bw_t$ for the convex combination, while $EGAB\mbox{-}P$ uses the uniform vector $\bu$. These thresholds determine the balance between the positive and negative elements in the resulting gradients, and due to the monotonicity of the deformed exponential transformation, this automatically controls the balance between portfolio entries of the weight vector that will experience a relative decrease or increase in their values. 
In our experience, despite the slight difference between thresholds, both gradient terms seem to perform quite similarly.

The most significant difference between the $EGAB\mbox{-}N$ and $EGAB\mbox{-}P$ is due to the second step of the updates, i.e., the normalizations of $\bw_{\star}$ shown in \eqref{Iter2-7} and \eqref{AB-EG-proj_iter-2}, i.e.,
\eq{
	\!\left\{
	\begin{array}{l}
		\displaystyle
	\bw_{t+1} =  \bw_{\star}/\|\bw_{\star}\|_1 \qquad \text{\small (normalization by scaling)}\\ 
	\bw_{t+1} = \text{\small projection onto the unit $l_1$-norm  simplex}(\bw_{\star})
	\end{array}
	\right.
}
On the one hand, as it can be seen in Listing 1, when $\|\bw_\star\|_1\leq 1$, both normalizations turn out to be equivalent. On the other hand, when $\|\bw_\star\|_1> 1$, they can only be regarded as equivalent in the neighbourhood of the uniform weight vector $\bw_{\star}\approx \bu$. While they tend to differ significantly for large learning rates $\eta$ or when sparsity of $\bw_{\star}$ increases. The geometry of the projections onto the feasible domain indicates that the normalization by scaling of $EGAB\mbox{-}N$ would generally correspond to smoother transitions, whereas, the minimum distance projection of $EGAB\mbox{-}P$ tends to promote the sparsity of the solutions for large values of $\eta$. 

The convergence analysis of the proposed algorithms falls outside the scope of this study. However, such analysis has been already done in \cite{Hill2001}--\!\cite{Orazio2021} for standard exponentiated gradient and mirror descent algorithms, which share some similarities with the proposed methods. 

\section{Illustrative example of application: gradient-based strategies for Online Portfolio Selection}
In this section, we derive the specific form of the EGAB updates for OnLine Portfolio Selection (OLPS). 
This is a fundamental research problem in area of computational finance \cite{Cover1991}-\cite{Tsai2023}, which has drawn extensive investigations in both machine learning and computational finance communities, especially for high frequency trading where is necessary to use relatively fast and robust algorithms. OLPS has become increasingly popular in recent years, particularly with the growth of online trading platforms and the availability of real-time market data.

The goal of OLPS is to maximize the return of an investment portfolio by dynamically allocating assets to different financial instruments. OLPS algorithms typically learn from past market data to adjust the portfolio allocation strategy over time. Therefore, the OLPS problem is a sequential decision-making problem in which an investor is an online decision-maker who updates the portfolio at the beginning of each period.

Let us consider a self-financed, discrete-time and no-margin/non-short investment
environment with $N$ assets for $T$ trading periods. The period can be
chosen arbitrarily, such as fraction of second, minute, hour, day, and even weeks. In our experiments we used days. In the $t$-th period, the performance of assets can be described by a vector of price
relatives, denoted by $\bx_t=[x_{1,t},x_{2,t},\ldots,x_{N,t}]^T \in \mathbf{R}_+^N $, where $x_{i,t}, \;\; (i=1,2,\ldots,N)$ is the closing price ($p_{i,t}$) of the $i$-th asset in the period $t$ divided by its closing price in the previous period ($p_{i,t-1}$), i.e. 
\eq{
	x_{i,t}=p_{i,t}/p_{i,t-1}
}

The portfolio, which reflects the investment decision in the $t$-th period, is
denoted by a weight vector $\bw_t= [w_{1,t},w_{2,t},\ldots,w_{N,t}]^T \in \mathbf{R}_+^N $
with the constraint
that  $w_{i,t}  \geq 0, \forall i,t$  and $\|\bw_t\|_1=1$.  
The $i$-th element
of $\bw_t$ specifies the proportion of the total portfolio wealth invested in $i$-th
asset in the $t$-th period.

We assume that the cumulative return obtained at the end of the $t$-th
period (e.g., one day) will be completely reinvested at the beginning of the $t+1$-th period and no additional wealth can be taken into the portfolio.
At the beginning, we assume that the portfolio is uniformly allocated, i.e.,
$\bw_0 = \bu\equiv \tfrac{1}{N}{\bf 1}$. On the $t$-th period, if the portfolio $\bw_t$ is adopted, when
the price relative vector $\bx_t$ occurs at the end of the $t$-th period, the wealth increases by a factor of $\bw_t^T \bx_t = \sum_{i=1}^N w_{i,t}\, x_{i,t}$. In the absence of transaction costs, the final cumulative wealth is given by the expression
\eq{
	CW_T^0 = CW_0 \; \prod_{t=1}^{T} \bw_t^T\bx_t 
	\label{CW_T}
}
where $CW_0$ denotes the initial wealth, which, for simplicity, can be set to 1 (e.g., one dollar of initial investment).

\subsection{Implementation of the EGAB updates for online portfolio selection}

During training and in absence of a relative-prices preprocessing, we estimate $\hat{\bx}_{t+1}$ by $\bx_{t}$, and define the opposite of the daily log-return as
\eq{
	L(\bw) = -\log(\bw^T \hat\bx_{t+1}). \label{crit-OLPS}
}
A variation of this negative log-return, denoted by $L_\Delta(\bw)$, will be the target loss function to minimize. Here, the sub-index $\Delta\in\{I,\parallel\}$ allows us to choose between the scale-invariant version of the loss function $L_I(\bw)= L(\bw/\|\bw\|_1)$, and its effective linear approximation $L_\parallel(\bw)$, which was defined in \eqref{Lparallel}. The regularization of this loss function by the AB~divergence yields the proposed criterion to optimize
\eq{
	J_t(\bw) = L_\Delta(\bw)
	+\lambda D^{(\alpha,\beta)}_{AB}(\bw\|\bw_{t}), 
}
subject to $\|\bw\|_1=1$ and $w_i>0\ \forall i$, being $\lambda =1/\eta$ a regularizer or penalty parameter which controls the smoothness of the solution. Differentiation of $L_\Delta(\bw_{t})$, with help of \eqref{Grad-Lpar}, lead us to the required gradients 
\eq{
	\nabla_{\bi w} L_\Delta(\bw_{t}) 
	=\left\{
	\begin{array}{ll}
	- \left(\hat\bx_{t+1}- \tilde{x}_{t+1} {\bf 1} \right)/\tilde{x}_{t+1} & \Delta=I
	\\
	- \left(\hat\bx_{t+1}- \bar{x}_{t+1} {\bf 1} \right)/\tilde{x}_{t+1}  & \Delta=\ \parallel
	\end{array}
	\right.
}
where $\bu\dot=\tfrac{1}{N}{\bf 1}$ denote the uniform portfolio and
\eq{
	\tilde{x}_{t+1}= \bw_t^T \hat\bx_{t+1},
	\qquad 
	\bar{x}_{t+1}= \bu^T\hat\bx_{t+1}.
}

Based on results of the previous sections and the above expressions we obtain, for $\Delta=I$, the $EGAB\mbox{-}N(\alpha,\beta)$ update 
\eq{
	\bw_{\star} &= \bw_{t} \odot \exp_{1-\beta}
	\left(\eta\
	\bw_{t}^{\gamma} \odot 
	\left(\hat\bx_{t+1}- \tilde{x}_{t+1} {\bf 1} \right)/\tilde{x}_{t+1}
	\right) 
	\label{Iter1-8}\\
	&= \left\{
	\begin{array}{ll}
		\displaystyle \bw_{t} \odot \left[1 - \beta \; \eta \; \bw_{t}^{\gamma} \odot 
		\nabla_{\bi w} L_I(\bw_{t})
		\right]_+^{1/\beta}, & \beta\neq 0,\\
		\bw_{t} \odot \exp \displaystyle(-\eta \; \bw_{t}^{\gamma} \odot \nabla_{\bi w} L_I(\bw_{t})), & \beta=0,
	\end{array}
	\right. \nonumber \\
	\bw_{t+1}   &=  \bw_{\star}/\|\bw_{\star}\|_1 .
	\label{Iter2-8}
}
where $\gamma=1-\alpha-\beta$.
Similarly, for $\Delta=\parallel\,$, we obtain the $EGAB\mbox{-}P(\alpha,\beta)$ update, which is implemented by
\eq{
	\bw_{\star} &= \bw_{t} \odot \exp_{1-\beta}
	\left(\eta\
	\bw_{t}^{\gamma} \odot 
	\left(\hat\bx_{t+1}- \bar{x}_{t+1} {\bf 1} \right)/\tilde{x}_{t+1}
	\right) \label{Iter1-9}\\
	\bw_{t+1}   &= \text{\small projection onto the unit $\ell_1$-norm simplex}(\bw_{\star}).
	\label{Iter2-9}
}

\color{black}
\subsection{Representative gradient based strategies for OnLine Portfolio Selection}\label{Strategies}

In our simulations, the proposed algorithms are compared with classical and state-of-the-art OLPS strategies such as the standard Exponentiated Gradient (EG) \cite{EG,Bouneffouf2016}, Passive-Aggressive Mean-Reversion (PAMR) \cite{Li2012}, Online Moving Average Reversion (OLMAR) \cite{Li2015}, and Robust Median-Reversion (RMR) \cite{RMR}. We also use the uniform buy and hold benchmark (UBAH) as a baseline strategy \cite{Li2014}. In the UBAH strategy, the portfolio starts being uniformly distributed over all the assets and then passively follows the dynamics of market movements throughout the trading period. 

These OLPS strategies can be historically classified into the categories of "follow the winner" (FTW) or momentum approaches, and "follow the loser" (FTL) or contrarian approaches. In general, OLPS algorithms can belong to of one of these categories depending both on the definition of their loss function and on the possible preprocessing of the relative prices. FTW strategies try to invest in assets that have shown a recent positive performance. Whereas, FTL strategies invest in the assets with poor recent performance in the hypothesis that they are underestimated by the market. 

Several FTL gradient-based techniques in the state-of-the-art exploit the mean market reversion hypothesis. Among them, we consider: PAMR, OLMAR, and RMR.

PAMR \cite{Li2012} is a strategy that sets a threshold for the mean market reversion over single trading periods. When the daily return of the portfolio is below the threshold, the algorithm retains the preceding portfolio, thereby facilitating a passive reversion to the mean. This aims to prevent mining stocks that may be potentially volatile. Conversely, if the daily return of the portfolio surpasses the threshold, the portfolio actively initiates a rebalancing process, targeting the most underperforming assets, in the belief that the relative stock prices will revert to the mean in the next trading day. 

OLMAR and RMR are strategies that exploit the more realistic hypothesis of a mean reversion over multiple trading periods. They both rely on the estimates $\hat{p}_{i,t+1}$ of the central location of the future price based on a given temporal window. While OLMAR considers online moving averages \cite{Li2015} for its estimates, which are potentially more sensitive to outliers, RMR considers robust adaptive median estimates \cite{RMR}. In the end, both implementations rely on a preprocessing for training of the input dataset, which replaces the true relative prices by the predicted ones as follows:
\eq{
	\hat{x}_{i,t+1}	\leftarrow \hat{p}_{i,t+1}/p_{i,t}.
}
 
Apart from this preprocessing, which for PAMR is implemented with the estimate $\hat{\bx}_{t+1}=\bx_{t}$, the three mean reversion techniques share the structural similarity of their proposed updates, which are given by
\eq{
	\bw_{\star} 
	&\equiv \bw_{t} +\eta_t' \ (\hat{\bx}_{t+1} - \bar{x}_{t+1} {\bf 1}).
	\label{Iter1-9cx}\\
	\bw_{t+1} &= \text{\small projection onto the unit $l_1$-norm simplex}(\bw_{\star}),
	\label{Iter2-9cx}
}
with learning rates that are determined in closed-form by
\eq{
	&\eta_t' = \frac{ s[s(\epsilon-\bw_t^T \hat\bx_{t+1})]_+} 
	{\| \hat\bx_{t+1} - \bar{x}_{t+1} {\bf 1} \|_2^2+\varepsilon}, 
	\\
	&\text{for }\
	s=\left\{
	\begin{array}{lc} 
		\!\!-1 &\!\! \text{\small PAMR}\\
		\!\!+1 &\!\! \text{\small OLMAR\ \&\ RMR},
	\end{array}
	\right. 
	\label{Iter0-9cx}
}
where $\varepsilon\rightarrow 0$ is an arbitrary negligible positive constant.

\subsection{Seeing PAMR, OLMAR and RMR through the lenses of generalized EGAB algorithms}

Consider the the OLPS loss function, the minus daily log-return in \eqref{crit-OLPS}, with the Euclidean divergence as regularizer (for which $\alpha=1$ and $\beta=1$). The optimization of the regularized function with the $EGAB\mbox{-}P(1,1)$ algorithm has $\gamma=-1$, resulting the update
\eq{
	\bw_{\star} 
	&= \bw_{t} \odot \exp_{0}
	\left(\eta\
	\bw_{t}^{-1} \odot (\hat\bx_{t+1} - \bar{x}_{t+1} {\bf 1})/\tilde{x}_{t+1}\right)
	\label{Iter2-9cx1}\\	
	&\equiv \bw_{t} +\eta'_t (\hat\bx_{t+1} - \bar{x}_{t+1} {\bf 1}),
	\label{Iter2-9cx2}
	\\
	\bw_{t+1}   &= \text{\small projection onto the unit $\ell_1$-norm simplex}(\bw_{\star}).
	\label{Iter2-9cx3}
}
where $\eta'_t=\eta/\tilde{x}_{t+1}$. 

In the case of the Euclidean divergence, not surprisingly, the multiplicative update in \eqref{Iter2-9cx1}-\eqref{Iter2-9cx2} turns out to coincide exactly with the additive Gradient Descent update in \eqref{Iter1-9cx}, which is at the core of the mean reversion techniques: PAMR, OLMAR and RMR. Whereas, later, the intermediate solution $\bw_{\star}$ is projected onto the feasible set. Therefore, their only significant difference with the $EGAB\mbox{-}P$ algorithm is that for the later we don't have a close-form solution for $\eta$, akin to~\eqref{Iter0-9cx}.

In general, $EGAB\mbox{-}P$ and $EGAB\mbox{-}N$ algorithms depend on three hyper-parameters ($\alpha$, $\beta$, and $\eta$) that provide us with extra degrees of flexibility to adapt the update to the particularities of the data distribution. For a good performance, we need to learn these hyper-parameters from preceding trading periods within each dataset.

\subsection{Taking into account transaction costs}

In the financial markets, transactions in the stock market invariably come with associated costs. In the linear model, these costs manifest as penalties proportional to the transactions size, with the commission rate $c_r$ acting as the factor of proportionality.

As is commonly understood,  the portfolio $\bw_t$ reflects the distribution of wealth among individual stocks, while the prices of the stocks typically undergo changes following a trading period. Consequently, at the conclusion of the trading period, the wealth allocation across stocks is adjusted, leading to an automatic modification of the resulting portfolio, which is then given by 
\eq{
	\bw_t' \equiv \frac{\bw_t\odot \bx_t}{\bw_t^T \bx_t} .
	\label{w_end} 	
}
Given the available information at time $t$, the algorithms return ${\bw}_{t+1}$, which represents the portfolio recommendation for the beginning of the next trading period. The size of the stock exchange from portfolio ${\bw}_t'$ to ${\bw}_{t+1}$ is referred as the turnover (see \cite{RMR})
\eq{
	T_t(\bw_{t+1}) \equiv \frac{1}{2}\|\bw_{t+1}-\bw_{t}'\|_1 .	
}
In a linear transaction cost model, the daily return after subtracting the commission of the transaction is given by
\eq{
	r_t = (\bw_t^T \bx_t)\ [1-c_r\ T_t(\bw_{t+1})].
}
and the cumulative wealth after $T$ trading periods is equal to
\eq{
	CW_T(c_r) =  CW_0 \prod_{t=0}^T  \left[(\bw_t^T \bx_t)\ (1-c_r\ T_t(\bw_{t+1}))\right].
}
Its negative logarithmic value decomposes as 
\eq{
	-\log CW_T(c_r) = L^{test}_t(\bw_{t+1}) + \text{const},
}
where the test loss function that we define as
\eq{
	L^{test}_t(\bw_{t+1}) 
	=  
	&-\log (\bw_{t+1}^T \bx_{t+1}) - \log (1-c_r\ T_t(\bw_{t+1})),\nonumber
}
captures all the dependence on $\bw_{t+1}$. Analogously, we define the surrogate training loss function $L_t^{train}(\bw)$ to optimize, where with respect to $L^{test}_t(\bw_{t+1})$ we replace the future relative-price $\bx_{t+1}$ by its estimate $\hat\bx_{t+1}$ depending on the chosen preprocessing option. We also add the extra training parameter $s\in\{\pm 1\}$ to choose between a FTL or a FTW strategy. In this way, we obtain the training loss function 
\eq{
	L_t^{train}(\bw) 
	=  
	-\log (\bw^T \hat\bx_{t+1})^s 
	-\log [1-c_r\ T_t(\bw)].
	\label{Ltrain1}
}
Now, we can optimize \eqref{Ltrain1} with respect to $\bw$ to determine the proposed portfolio $\bw_{t+1}$ for the next trading period. 

At this point, the implementation of the proposed algorithms simply requires the determination of the subgradient of the training loss at $\bw_t$. For positive weights, this subgradient is unique and coincident with  
\eq{
	\nabla_{\bi w} L_t^{train}(\bw_{t}) 
	&=  - s\frac{\hat\bx_{t+1}}{\bw^T \hat\bx_{t+1}} 
	+\frac{\mathrm{sign}(\bw-\bw'_t)}{\frac{2}{c_r}-\|\bw-\bw'_t\|_1}, 
}
where the adjusted portfolio $\bw_t'$ was defined in \eqref{w_end}. Note that the second term on the left-hand-size accounts for the existence of transaction costs.

The scale-invariant version of the loss function is defined by $L_I^{train}(\bw_{t})=L_t^{train}(\bw_{t}/\|\bw_{t}\|_1)$, and its gradient given by
\eq{
	\nabla_{\bi w} L_I^{train}(\bw_{t})=   \nabla_{\bi w} L_t^{train}(\bw_t)- (\bw_t^T \nabla_{\bi w} L_t^{train}(\bw_t))\, {\bf 1} \, .
	\label{eq_gradLIb}
}
The gradient component parallel to $\ell_1$ manifold is equal to
\eq{
	\nabla_{\bi w} L_\parallel^{train}(\bw_t) &=  \nabla_{\bi w} L_t^{train}(\bw_t)- (\bu^T \nabla_{\bi w} L_t^{train}(\bw_t))\, {\bf 1}
	\label{Grad-Lparb}\, .
}
Finally, these gradient expressions are respectively substituted in \eqref{Iter1-7} and \eqref{AB-EG-proj_iter-1} to implement $EGAB\mbox{-}N(\alpha,\beta)$ and  $EGAB\mbox{-}P(\alpha,\beta)$ updates that take into account the existence of transaction costs.

\section{Computer Simulation Experiments}
To assess the validity and performance of the EGAB updates in the context of Online Portfolio Selection (OLPS), we conducted experiments using real-life financial historical data. A total of sixteen diverse datasets were employed for this purpose, which details are provided in Table~\ref{Datasets}. 

The first four datasets (NYSE-N, NYSE-O, MSCI, and TSE) are commonly utilized in OLPS literature for evaluation, as documented in references \cite{Li2014}-\cite{Li2023}, and are available in the OLPS toolbox \cite{OLPS}. Readers interested in a comprehensive description of these datasets are directed to section 5.1 and table~3 of \cite{Li2012}. 

Another dataset under consideration is the historical dataset of the Dow Jones Industrial Average (DOW30), which tracks the top 30 biggest US companies, primarily in the industrial sector. This dataset spans from the beginning of January 2005 to the end of December 2021, encompassing a total of $T=4280$ trading days. Notably, it covers significant market events, including the subprime mortgage crisis from 2007 to 2009 and the onset of the COVID-19 pandemic in March 2020. 

Additionally, we enriched the dataset collection by incorporating specific sectors of the Standard \& Poor's 500 index. These sectors include: information technology (SP-IT), consumer staples (SP-CS), communication services (SP-CSe), financials (SP-F), industrials (SP-I), real estate (SP-RE), energy (SP-E), healthcare (SP-HC),
materials (SP-M), and Utilities (SP-U). 

\begin{table*}
	\color{black}
	\caption{Summary of the considered datasets}
	\label{Datasets}
	\centering
	\vspace{0.5cm}  
	\begin{tabular}{|l|c|c|c|c|c|c|c|c|c|c|c|c|c|c|}
		\hline
		\bf Dataset & \bf Acronym & \bf Time Frame & \bf \#\,Days & \bf \#\,Assets 
		\\
		\hline\hline
		New York Stock Exchange (N) & NYSE-N & 01/01/1985 - 30/06/2010 & 6431 & 23 
		\\\hline
		New York Stock Exchange (O) & NYSE-O & 03/07/1962 - 31/12/1962 & 5651 & 36 
		\\\hline
		MSCI World Index& MSCI &01/04/2006 - 31/03/2010 & 1043 & 24 
		\\\hline   
		Toronto Stock Exchange & TSE & 04/01/1994 -  31/12/1998 & 1259 & 88 
		\\\hline \hline 
		Dow Jones Industrial Average & DOW30 & 31/12/2004  - 31/12/2021& 4280 &30 
		\\\hline 
		S\&P 500 Industrial Technology & SP-IT &31/12/2004  - 31/12/2021& 4280 & 52 
		\\\hline
		S\&P 500 Consumer Staples & SP-CS & 31/12/2004  - 31/12/2021 & 4280 & 55 
		\\\hline
		S\&P 500 Communication Services & SP-CSe &  31/12/2004  - 31/12/2021& 4280 & 13 
		\\\hline
		S\&P 500 Financials & SP-F &31/12/2004  - 31/12/2021& 4280 & 55 
		\\\hline
		S\&P 500 Industrials & SP-I &  31/12/2004  - 31/12/2021& 4280 & 55 
		\\\hline
		S\&P 500 Real Estate & SP-RE & 31/12/2004  - 31/12/2021& 4280 & 29 
		\\\hline
		S\&P 500 Energy & SP-E & 31/12/2004  - 31/12/2021 & 4280 & 32 
		\\\hline
		S\&P 500 Health Care & SP-HC & 31/12/2004  - 31/12/2021& 4280 & 57 
		\\\hline
		S\&P 500 Materials & SP-M &  31/12/2004  - 31/12/2021& 4280 & 21 
		\\\hline
		S\&P 500 Utilities & SP-U &  31/12/2004  - 31/12/2021 & 4280 & 27 
		\\\hline

	\end{tabular}
\end{table*}

\subsection{Comparison of gradient-based algorithms}
The proposed comparison is aimed to corroborate through simulations the following key contributions. Firstly, that the generalizations we propose of exponentiated gradient algorithms, based on the AB divergence, allows for a beneficial adaptation to dataset distributions through the learning of the algorithm hyperparameters: $\alpha$, $\beta$,  $\eta$ and of the data preprocessing option. This will be validated through a resulting  improvement of the financial indicators. Secondly, to corroborate that the optimization of the proposed criterion in \eqref{Ltrain1}, which accounts for the existence of transaction costs, also leads to a significant improvement in performance through the progressive attenuation of the turnover with the increase of these costs.

The simulations compare the proposed EGAB algorithms with the representative OLPS strategies based on gradient iterations: PAMR, OLMAR, RMR, and standard EG. Additionally, we consider the uniform buy and hold benchmark (UBAH) as a baseline strategy. Those portfolio selection algorithms that outperform UBAH are said to ``beat the market''. 

In the literature, the criteria optimized by all of these representative strategies do not account for the existence of transaction costs. Their algorithmic configurations for the simulations have been adopted from the recommendations given by Section 5.3 of \cite{RMR}. In particular, they advise for EG to use $\eta=0.05$, for PAMR $\epsilon = 0.5$, and for OLMAR and RMR $\epsilon = 5$ together with their respective estimation of mean and median prices $\hat{p}_{i,t+1}$ through a sliding temporal window of 5 samples. 

For the EGAB algorithms, apart from their default parameters, we consider the choice of preprocessing for the data and a training strategy. As it is well known, the direct tuning of the algorithmic parameters, leads to an overfit to the data and to an overestimation of the predictive performance. To prevent this effect, we split each dataset of length $T$ into an initial validation set of $T_{val}=\tfrac{1}{8}T$ samples, and a test set of $T_{test}=\tfrac{7}{8}T$ samples. So we will run the algorithms on the initial trading periods with different hyperparameter configurations to learn the choices that leads to the best validation performance. The algorithms with their learned configurations are later run on the test set to obtain their predictive performance.  

We evaluate in simulations the proposed exponential gradient algorithms: $EGAB\mbox{-}N$ and $EGAB\mbox{-}P$, together with a third one $EG+$ that will be introduced below. These three proposals use the validation dataset to learn the best hyperparameters and they optimize the proposed criterion in~\eqref{Ltrain1}, which accounts for the possible existence of transaction costs. They learn the regularization weighting term from a prefixed set of possibilities $\lambda\in\{2^{-10},2^{-9},\ldots,2^0,2^{1}\}$, as well as $s\in\{\pm1\}$, a switching parameter that maintains (for $s=1$) or reverses for ($s=-1$) the default algorithm strategy, commuting between FTW and FTL alternatives. They also learn the best preprocessing option for the relative prices withing the set of alternatives
\eq{
	\hat{\bx}_{t+1}=
	\left\{
	\begin{array}{l}
		\bx_{t} \\
		\text{mean}(\bp_t,\ldots,\bp_{t-n})\oslash \bp_{t} \\
		\ell_1\text{-median}(\bp_t,\ldots,\bp_{t-n})\oslash \bp_{t}
	\end{array}
	\right.
}
where $\oslash$ refers to the component-wise division of the vector elements. Options that coincide with the respective preprocessing choices of PARM \cite{Li2012}, OLMAR \cite{Li2015} and RMR \cite{RMR}. 

$EG+$ will be used to refer to classical EG algorithm for the optimization of~\eqref{Ltrain1}, together with the learning of: the regularizer parameter $\lambda=\eta^{-1}$ (see \cite{EGMA} for a similar learning proposal), the preprocessing option for $\hat{\bx}_{t+1}$, as well as the considered strategy $s\in\{\pm1\}$. The "$+$" symbol of $EG+$ is a simply reminder that it is basically the classical EG algorithm with an optimized or learned configuration. Similarly, the hyperparameters $(\alpha,\beta)$ of EGAB algorithms are learned within the subset of representative candidate tuples $\{(1,1),(1,1/2),(5,-5)\}$. 

The predictive performance of the algorithms is evaluated over a disjoint test set,  which comprises the last $T_{test}$ trading periods of each dataset. 
Table~\ref{table-CW} presents the cumulative wealth  $CW_{T_{test}}(c_r)$ obtained by the algorithms for several commission rates $c_r\in\{0\%, 0.025\%,0.25\%,0.1\%,0.25\%\}$.

Because of the high variability of this performance index, which is extreme for datasets such as NYSE-N and NYSE-O, the arithmetic mean does not constitute a good predictor of performance, while the geometric mean provides a conservative estimator which mitigates the influence of extreme results. Therefore, in the table, we opted for presenting the geometric mean of the cumulative wealth, which is tantamount to presenting in natural units the result of evaluating the mean of the cumulative wealth in dB.

In absence of commission rates, one can observe in Table~\ref{table-CW}, that the gradient-based algorithms usually beat the market (the baseline result of UBAH). In general, the best cumulative wealth clearly depends on the algorithm and dataset, without a clear winner. However, the $EGAB\mbox{-}P$ proposal provides a good geometric mean performance over the datasets, which is closely followed in descending ranking by RMR, OLMAR, $EGAB\mbox{-}N$ and PAMR. 

The comparison between EG and $EG+$, reveals that learning of the algorithm configuration clearly boosts its performance. Note that $EG+$ coincides with the particularization of $EGAB\mbox{-}N$ for $(\alpha,\beta)=(1,0)$. Besides, the freedom in learning $(\beta,\gamma)$ in this later algorithm, allows a certain adaptation to the distribution of the data, which seems to improve its results with respect to those obtained by $EG+$. However, the best performance is obtained for $EGAB\mbox{-}P$, because the simplex projection of the updated weights enhances the sparsity in comparison with the scalar normalization used by $EGAB\mbox{-}N$.
 
\begin{figure}[t]
	\begin{center}
		\includegraphics[trim = 0mm 0mm 0mm 0mm, clip,width=.6\linewidth] {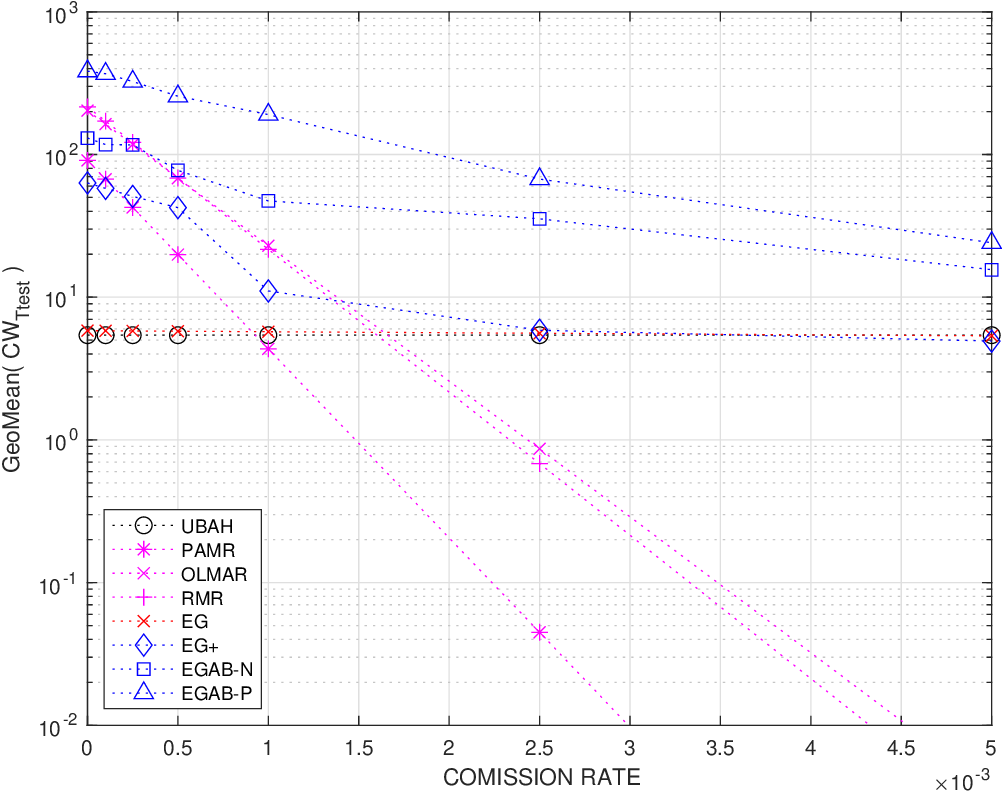}	
		\caption{Illustration of the decrease in geometric mean of cumulative wealth (over all the datasets) with the increase of the commission rate.}
		\label{Fig_turnover}	
	\end{center}
\end{figure}

\begin{figure}[t]
	\begin{center}
		\includegraphics[trim = 14mm 0mm 0mm 7mm, clip,width=.7\linewidth] {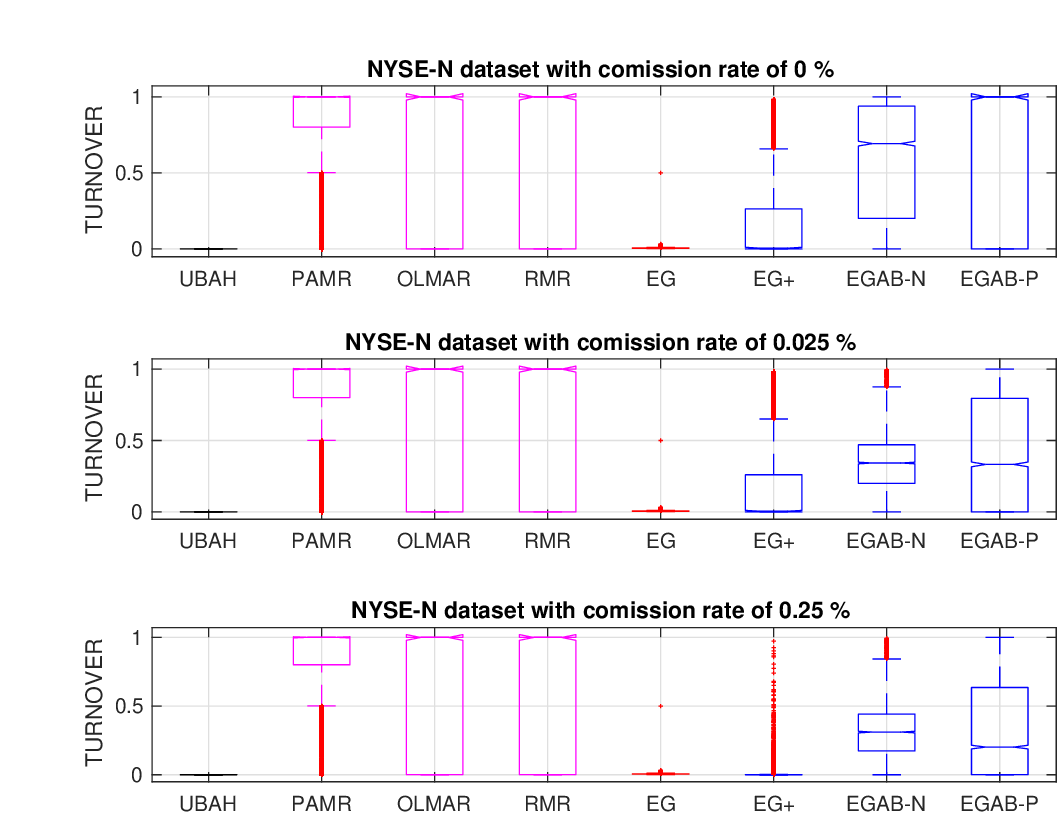}	
		\caption{Boxplots of the turnover of the algorithms, for the NYSE-N dataset, when varying the commission rate $c_r\in\{0\%,0.025\%,0.25\%\}$. 
		The figure illustrates how the proposed algorithms (in blue) progressively lower their turnover when commissions rise.}
		\label{Fig_boxplot}	
	\end{center}
\end{figure}

Figure \ref{Fig_turnover} illustrates how the expected cumulative wealth of the algorithms is affected by the commission rate. Clearly, as the rate increases, most of the techniques degrade their results, except for UBAH, which by design has zero turnover and does not incur in transaction costs. The cumulative wealth degradation due to these costs is quite severe for PAMR, OLMAR, and RMR. From these results one can observe that the EGAB algorithms  exhibit a greater resilience with respect to the increase of commissions, mainly, because their optimized criteria already have them into account. 
Figure \ref{Fig_boxplot} represents the evolution of turnover statistics for NYSE-N dataset when $c_r\in\{0\%,0.025\%,0.25\%\}$. One can observe how the $EG+$, $EGAB\mbox{-}N$, and $EGAB\mbox{-}P$ algorithms progressively lower their turnover as the transaction costs increase for higher commission rates. 

For comparison purposes with the existing literature, it may be useful to have some insight of the cumulative wealth performance ${CW}_{T}(c_r)$ over the whole extension of the dataset, i.e., $T$ samples. However, one has to be cautious with its computation. Directly evaluating it would lead to overfitting the data, as we have used $T_{val}$ of the samples to train the hyperparameters. 
Instead, a meaningful extrapolation of $CW_{T}(c_r)$ from  $CW_{T_{test}}(c_r)$ can be always obtained through the simple expression
\eq{
	\widehat{CW}_{T}(c_r)=(CW_{T_{test}}(c_r))^{T/T_{test}}.
	\label{extrapolation}
}

For the sake of completeness, the performance of the compared methods is also evaluated in terms of the following financial metrics: Sharpe ratio (SR), Calmar ratio (CR), Annualized Percentage Yield (APY), and maximum drawdown (MDD). The APY, which represents the profit or effective rate of return for one year. MMD is an indicator of the downside risk of the investment. SR measures its risk-adjusted return where the adjusted performance is compared against the return volatility. The CR ratio measures the ratio between the profit performance of an investment and its maximum drawdown risk. Although higher values of APY, SR, CR together with lower values of MDD result preferable, in general, one usually have to reach a compromise or balance between them.

In Table~\ref{Tab_financial_measures1} these indicators are evaluated in absence of  transactions costs, i.e. for $c_r=0\%$. There results show that EGAB together with OLMAR and RMR, all perform quite well in terms of APY, SR and CR. On the contrary, their MDD index is usually much higher than the conservative approaches based on UBAH and EG, whose profits are quite low. The results also reveal a clear distinction between the group of the first four datasets (NYSE-N, NYSE-O, MSCI and TSE) with the remaining ones. For the first group, most of the financial indicators are much better than for the second, which is a much greater challenge for all algorithms.
The financial indicators have been also evaluated in presence of transaction costs. Table~\ref{Tab_financial_measures2} presents them for a moderate commission rate of $c_r=0.025\%$, and Table~\ref{Tab_financial_measures3} for a higher rate of $c_r=0.1\%$. The results of these tables confirm a progressive degradation of all indicators with higher commissions, being quite significant for PAMR, OLMAR and RMR, and much more resilient for the $EGAB\mbox{-}P$ algorithmic proposal.

\begin{table*}
	\caption{Detailed values of the $CW_{T_{test}}(c_r)$ of the considered OLPS methods for several commission rates $c_r$. The cumulative wealth is evaluated only over the test set, which comprises the last $T_{test}=\frac{7}{8}T$ trading periods of the dataset.}
	\label{table-CW}
	\centering
	\color{black}
	\begin{tabular}{|c|c|c|c|c|c|c|c|c|c|}
	\hline
	\multirow{3}{*}{$\boldsymbol{c_r}$} 
	& \multirow{3}{*}{\bf{Dataset}}  & 
	\multicolumn{8}{c|}{\bf{$\boldsymbol{CW_{T_{test}}(c_r)}$\ \ for each dataset}}\\ \cline{3-10}
	& & \multicolumn{5}{c|}{\bf{Representative OLPS methods}} & \multicolumn{3}{c|}{\bf{EGAB proposals}} \\
	\cline{3-10}
	& & \bf{UBAH} 
	& \bf{PAMR} & \bf{OLMAR} & \bf{RMR} & \bf{EG} & $\boldsymbol{EG+}$ & $\boldsymbol{EGAB\mbox{-}N}$ & $\boldsymbol{EGAB\mbox{-}P}$ 
	\\
	\hline\hline
	
\multirow{15}{*}{\rotatebox[origin=c]{90}{\bf 0.00 \%}} 
& NYSE-N  & 8.68 & 1.58e+05 &  \bf 1.94e+07 & 1.76e+07 & 15.28 & 4.78e+05 & 5.80e+06 & 1.76e+07 \\\cline{2-10}
& NYSE-O  & 8.86 & 1.90e+13 & 1.78e+14 & 3.31e+14 & 13.68 & 3.72e+06 & 5.56e+12 &  \bf 3.90e+15 \\\cline{2-10}
& MSCI    & 0.88 & 12.63 & 11.53 &  \bf 14.62 & 0.89 & 3.60 & 6.57 & 11.60 \\\cline{2-10}
& TSE     & 1.67 & 107.05 & 14.15 & 32.25 & 1.59 & 35.49 & 77.28 &  \bf 126.52 \\\cline{2-10}
& DOW30   & 7.30 & 0.29 & 3.00 & 3.34 & 6.18 &  \bf 7.33 & 2.31 & 4.19 \\\cline{2-10}
& SP-IT   & 12.23 & 12.23 & 5.85 & 4.02 & 13.88 & 39.25 & 28.98 &  \bf 139.64 \\\cline{2-10}
& SP-CS   & 5.74 & 6.30 &  \bf 9.28 & 8.31 & 6.04 & 2.65 & 4.19 & 4.47 \\\cline{2-10}
& SP-CSe  & 19.18 & 4.22 & 14.67 & 12.85 & 9.23 & 9.79 & 1.00 &  \bf 202.19 \\\cline{2-10}
& SP-F    & 4.14 & 1.94 & 46.44 & 20.40 & 4.66 &  \bf 60.12 & 7.88 & 7.69 \\\cline{2-10}
& SP-I    & 7.48 & 8.49 & 7.26 & 17.99 & 8.47 &  \bf 131.37 & 19.29 & 12.15 \\\cline{2-10}
& SP-RE   & 4.89 & 7.40 & 15.48 & 12.31 & 5.10 & 11.41 &  \bf 34.70 & 4.38 \\\cline{2-10}
& SP-E    & 3.09 & 1.05 & 6.21 & 5.54 & 3.66 & 1.89 & 6.66 &  \bf 16.13 \\\cline{2-10}
& SP-HC   & 9.18 & 13.80 & 2.52 & 5.64 & 10.30 &  \bf 151.99 & 1.51 & 5.43 \\\cline{2-10}
& SP-M    & 5.60 & 65.36 &  \bf 78.90 & 70.20 & 7.10 & 6.76 & 32.76 & 56.33 \\\cline{2-10}
& SP-U    & 3.81 & 5.42 & 4.68 & 2.59 & 3.94 & 3.44 &  \bf 6.77 & 5.32 \\\cline{2-10}
\hline   
\multirow{1}{*}{{\bf{Geometric mean }}}
& -       & 5.42 & 90.90 & 202.91 & 215.59 & 5.82 & 63.18 & 130.48 &  \bf 383.56 \\ \hline\hline

\multirow{15}{*}{\rotatebox[origin=c]{90}{\bf 0.025 \%}} 
& NYSE-N  & 8.68 & 4.86e+04 &  \bf 8.31e+06 & 7.25e+06 & 15.16 & 4.57e+05 & 2.51e+06 & 6.10e+06 \\\cline{2-10}
& NYSE-O  & 8.86 & 6.67e+12 & 7.91e+13 & 1.42e+14 & 13.58 & 2.52e+06 & 2.55e+12 &  \bf 2.09e+15 \\\cline{2-10}
& MSCI    & 0.88 & 10.31 & 9.90 &  \bf 12.46 & 0.89 & 3.48 & 5.93 & 10.12 \\\cline{2-10}
& TSE     & 1.67 & 86.70 & 12.01 & 27.12 & 1.59 & 17.66 & 64.20 &  \bf 101.04 \\\cline{2-10}
& DOW30   & 7.30 & 0.13 & 1.69 & 1.82 & 6.16 & 6.61 &  \bf 49.34 & 2.30 \\\cline{2-10}
& SP-IT   & 12.23 & 5.63 & 3.40 & 2.26 & 13.82 & 33.50 & 24.24 &  \bf 134.74 \\\cline{2-10}
& SP-CS   & 5.74 & 2.80 & 5.30 & 4.61 &  \bf 6.02 & 2.46 & 4.19 & 4.47 \\\cline{2-10}
& SP-CSe  & 19.18 & 1.95 & 8.77 & 7.21 & 9.19 & 4.65 & 0.99 &  \bf 202.09 \\\cline{2-10}
& SP-F    & 4.14 & 0.87 & 26.06 & 11.06 & 4.64 &  \bf 46.70 & 7.88 & 7.69 \\\cline{2-10}
& SP-I    & 7.48 & 3.81 & 4.10 & 9.82 & 8.44 &  \bf 125.64 & 9.91 & 6.87 \\\cline{2-10}
& SP-RE   & 4.89 & 3.29 & 8.58 & 6.68 & 5.08 & 9.95 & 23.13 &  \bf 23.54 \\\cline{2-10}
& SP-E    & 3.09 & 0.48 & 3.47 & 3.01 & 3.64 & 1.66 & 5.42 &  \bf 13.62 \\\cline{2-10}
& SP-HC   & 9.18 & 6.31 & 1.45 & 3.13 & 10.25 &  \bf 136.51 & 0.87 & 4.62 \\\cline{2-10}
& SP-M    & 5.60 & 29.31 &  \bf 45.07 & 39.12 & 7.06 & 5.44 & 15.74 & 24.59 \\\cline{2-10}
& SP-U    & 3.81 & 2.40 & 2.65 & 1.43 & 3.93 & 3.02 & 6.30 &  \bf 6.71 \\\cline{2-10}
\hline   
\multirow{1}{*}{{\bf{Geometric mean }}}
& -       & 5.42 & 42.53 & 117.68 & 121.36 & 5.80 & 50.82 & 116.76 &  \bf 324.79 \\ \hline\hline
	
\multirow{15}{*}{\rotatebox[origin=c]{90}{\bf 0.1 \%}} 
& NYSE-N  & 8.68 & 1.42e+03 & 6.49e+05 & 5.06e+05 & 14.79 & 1.17 & 5.86e+05 &  \bf 1.11e+06 \\\cline{2-10}
& NYSE-O  & 8.86 & 2.89e+11 & 6.90e+12 & 1.13e+13 & 13.30 & 1.04e+06 & 2.64e+11 &  \bf 3.46e+14 \\\cline{2-10}
& MSCI    & 0.88 & 5.62 & 6.27 &  \bf 7.70 & 0.89 & 0.46 & 5.11 & 6.94 \\\cline{2-10}
& TSE     & 1.67 & 46.04 & 7.34 & 16.12 & 1.58 & 7.10 & 36.75 &  \bf 49.38 \\\cline{2-10}
& DOW30   &  \bf 7.30 & 0.01 & 0.30 & 0.30 & 6.09 & 4.10 & 2.73 & 2.72 \\\cline{2-10}
& SP-IT   & 12.23 & 0.55 & 0.66 & 0.40 & 13.63 & 21.11 & 14.20 &  \bf 83.14 \\\cline{2-10}
& SP-CS   & 5.74 & 0.25 & 0.99 & 0.79 & 5.96 & 2.01 & 4.19 &  \bf 13.17 \\\cline{2-10}
& SP-CSe  & 19.18 & 0.19 & 1.87 & 1.27 & 9.06 & 8.46 & 0.99 &  \bf 201.81 \\\cline{2-10}
& SP-F    & 4.14 & 0.08 & 4.60 & 1.76 & 4.58 &  \bf 26.19 & 7.85 & 7.85 \\\cline{2-10}
& SP-I    & 7.48 & 0.34 & 0.74 & 1.59 & 8.34 & 2.96 &  \bf 8.78 & 5.29 \\\cline{2-10}
& SP-RE   & 4.89 & 0.29 & 1.46 & 1.07 & 5.02 & 7.86 & 8.70 &  \bf 11.83 \\\cline{2-10}
& SP-E    & 3.09 & 0.04 & 0.61 & 0.48 & 3.59 & 1.03 & 2.15 &  \bf 6.65 \\\cline{2-10}
& SP-HC   & 9.18 & 0.60 & 0.28 & 0.54 & 10.11 &  \bf 102.20 & 0.17 & 1.91 \\\cline{2-10}
& SP-M    & 5.60 & 2.64 & 8.39 & 6.76 & 6.97 & 5.19 & 2.09 &  \bf 11.08 \\\cline{2-10}
& SP-U    & 3.81 & 0.21 & 0.48 & 0.24 & 3.90 & 2.34 &  \bf 5.94 & 2.83 \\\cline{2-10}
\hline   
\multirow{1}{*}{{\bf{Geometric mean }}}
& -       & 5.42 & 4.27 & 22.95 & 21.66 & 5.73 & 11.06 & 47.17 &  \bf 190.26 \\ \hline\hline

\multirow{15}{*}{\rotatebox[origin=c]{90}{\bf 0.25 \%}} 
& NYSE-N  & 8.68 & 1.18 & 3.92e+03 & 2.44e+03 & 14.08 & 2.45 & 4.01e+04 &  \bf 1.55e+05 \\\cline{2-10}
& NYSE-O  & 8.86 & 5.37e+08 & 5.19e+10 & 7.11e+10 & 12.76 & 1.07e+04 & 1.09e+10 &  \bf 5.06e+12 \\\cline{2-10}
& MSCI    & 0.88 & 1.66 & 2.51 & 2.94 & 0.88 & 0.42 & 2.77 &  \bf 3.83 \\\cline{2-10}
& TSE     & 1.67 & 12.95 & 2.74 & 5.68 & 1.56 & 2.01 &  \bf 13.54 & 12.76 \\\cline{2-10}
& DOW30   &  \bf 7.30 & 0.00 & 0.01 & 0.01 & 5.96 & 1.70 & 2.72 & 2.71 \\\cline{2-10}
& SP-IT   & 12.23 & 0.01 & 0.02 & 0.01 & 13.25 &  \bf 45.66 & 14.17 & 14.08 \\\cline{2-10}
& SP-CS   & 5.74 & 0.00 & 0.03 & 0.02 & 5.85 & 0.96 & 4.18 &  \bf 13.13 \\\cline{2-10}
& SP-CSe  &  \bf 19.18 & 0.00 & 0.08 & 0.04 & 8.80 & 8.43 & 0.99 & 2.56 \\\cline{2-10}
& SP-F    & 4.14 & 0.00 & 0.14 & 0.04 & 4.46 & 6.88 &  \bf 8.10 & 7.83 \\\cline{2-10}
& SP-I    & 7.48 & 0.00 & 0.02 & 0.04 &  \bf 8.15 & 1.66 & 5.24 & 5.71 \\\cline{2-10}
& SP-RE   & 4.89 & 0.00 & 0.04 & 0.03 & 4.91 & 3.43 & 3.09 &  \bf 6.99 \\\cline{2-10}
& SP-E    & 3.09 & 0.00 & 0.02 & 0.01 &  \bf 3.49 & 3.32 & 1.48 & 3.13 \\\cline{2-10}
& SP-HC   & 9.18 & 0.01 & 0.01 & 0.02 & 9.85 &  \bf 60.99 & 27.67 & 0.83 \\\cline{2-10}
& SP-M    & 5.60 & 0.02 & 0.29 & 0.20 & 6.78 & 0.91 & 2.09 &  \bf 11.06 \\\cline{2-10}
& SP-U    & 3.81 & 0.00 & 0.02 & 0.01 & 3.84 & 3.34 &  \bf 5.93 &  \bf 5.93 \\\cline{2-10}
\hline   
\multirow{1}{*}{{\bf{Geometric mean }}}
& -       & 5.42 & 0.00 & 0.85 & 0.70 & 5.58 & 5.87 & 35.43 &  \bf 67.29 \\ \hline
	
\end{tabular}
\end{table*}

\begin{table*}[t] 
	\caption{ Sharpe ratio ($SR$), Calmar ratio ($CR$), and annualized percentage yield ($APY$) for several  datasets for different OLPS methods, without trading commission rates ($c_r=0.0\%$). The best values of each index are highlighted in bold type.}
	
	\label{Tab_financial_measures1}
	\centering
	\color{black}
	\begin{tabular}{|c|c|c|c|c|c|c|c|c|c|}
	\hline
	\multirow{2}{*}{$\boldsymbol{Metrics}\ (c_r=0\%)$} & \multirow{2}{*}{\bf{Dataset}}  & \multicolumn{5}{c|}{\bf{Representative OLPS methods}} & \multicolumn{3}{c|}{\bf{EGAB proposals}} \\
	\cline{3-10}
	& & \bf{UBAH} 
	& \bf{PAMR} & \bf{OLMAR} & \bf{RMR} & \bf{EG} & $\boldsymbol{EG+}$ & $\boldsymbol{EGAB\mbox{-}N}$ & $\boldsymbol{EGAB\mbox{-}P}$ 
	\\
	\hline\hline
	
\multirow{15}{*}{\rotatebox[origin=c]{90}{\bf{Annualized Percentage Yield (\%)}}} 

& NYSE-N  & 10.20 & 70.90 &  \bf 112.00 & 111.00 & 13.00 & 79.60 & 100.80 & 111.00 \\\cline{2-10}
& NYSE-O  & 11.80 & 374.80 & 432.20 & 449.30 & 14.30 & 116.10 & 346.00 &  \bf 522.80 \\\cline{2-10}
& MSCI    & -3.40 & 101.20 & 96.20 &  \bf 109.50 & -3.10 & 42.30 & 68.00 & 96.60 \\\cline{2-10}
& TSE     & 12.50 & 190.90 & 83.20 & 121.10 & 11.20 & 126.00 & 170.00 &  \bf 202.20 \\\cline{2-10}
& DOW30   &  \bf 14.30 & -7.90 & 7.70 & 8.50 & 13.00 &  \bf 14.30 & 5.80 & 10.10 \\\cline{2-10}
& SP-IT   & 18.30 & 18.30 & 12.60 & 9.80 & 19.40 & 28.00 & 25.40 &  \bf 39.40 \\\cline{2-10}
& SP-CS   & 12.50 & 13.20 &  \bf 16.20 & 15.30 & 12.90 & 6.80 & 10.10 & 10.60 \\\cline{2-10}
& SP-CSe  & 22.00 & 10.20 & 19.80 & 18.70 & 16.10 & 16.60 & 0.00 &  \bf 42.90 \\\cline{2-10}
& SP-F    & 10.00 & 4.50 & 29.50 & 22.50 & 10.90 &  \bf 31.70 & 14.90 & 14.70 \\\cline{2-10}
& SP-I    & 14.50 & 15.50 & 14.30 & 21.50 & 15.50 &  \bf 38.80 & 22.00 & 18.30 \\\cline{2-10}
& SP-RE   & 11.30 & 14.40 & 20.20 & 18.40 & 11.60 & 17.80 &  \bf 26.90 & 10.50 \\\cline{2-10}
& SP-E    & 7.90 & 0.30 & 13.10 & 12.20 & 9.10 & 4.40 & 13.60 &  \bf 20.60 \\\cline{2-10}
& SP-HC   & 16.10 & 19.30 & 6.40 & 12.30 & 17.00 &  \bf 40.20 & 2.80 & 12.00 \\\cline{2-10}
& SP-M    & 12.30 & 32.50 &  \bf 34.20 & 33.10 & 14.10 & 13.70 & 26.50 & 31.20 \\\cline{2-10}
& SP-U    & 9.40 & 12.00 & 10.90 & 6.60 & 9.70 & 8.70 &  \bf 13.70 & 11.90 \\\cline{2-10}
\hline   
\multirow{1}{*}{{\bf{Mean APY (\%)}}}
& -       & 11.98 & 58.01 & 60.57 & 64.65 & 12.31 & 39.00 & 56.43 &  \bf 76.99 \\ \hline\hline
	
\multirow{15}{*}{\rotatebox[origin=c]{90}{\bf Sharpe Ratio}} 
& NYSE-N  & 0.35 & 1.35 &  \bf 1.86 & 1.85 & 0.48 & 1.34 & 1.79 & 1.85 \\\cline{2-10}
& NYSE-O  & 0.50 & 7.31 & 7.70 & 7.93 & 0.74 & 2.16 & 6.93 &  \bf 9.27 \\\cline{2-10}
& MSCI    & -0.29 & 2.55 & 2.25 &  \bf 2.59 & -0.27 & 0.91 & 1.79 & 2.29 \\\cline{2-10}
& TSE     & 0.65 &  \bf 2.53 & 0.82 & 1.23 & 0.55 & 1.70 & 1.93 & 2.02 \\\cline{2-10}
& DOW30   &  \bf 0.53 & -0.33 & 0.09 & 0.12 & 0.45 & 0.27 & 0.05 & 0.18 \\\cline{2-10}
& SP-IT   & 0.59 & 0.35 & 0.19 & 0.13 & 0.63 & 0.51 & 0.58 &  \bf 0.69 \\\cline{2-10}
& SP-CS   & 0.55 & 0.30 & 0.36 & 0.34 &  \bf 0.58 & 0.09 & 0.22 & 0.30 \\\cline{2-10}
& SP-CSe  & 0.62 & 0.15 & 0.37 & 0.35 & 0.54 & 0.31 & -0.10 &  \bf 0.76 \\\cline{2-10}
& SP-F    & 0.22 & 0.01 & 0.37 & 0.27 & 0.22 &  \bf 0.42 & 0.30 & 0.29 \\\cline{2-10}
& SP-I    & 0.45 & 0.30 & 0.24 & 0.40 & 0.49 &  \bf 0.81 & 0.51 & 0.33 \\\cline{2-10}
& SP-RE   & 0.25 & 0.23 & 0.34 & 0.30 & 0.24 & 0.30 &  \bf 0.51 & 0.14 \\\cline{2-10}
& SP-E    & 0.16 & -0.08 & 0.19 & 0.17 & 0.20 & 0.01 & 0.22 &  \bf 0.37 \\\cline{2-10}
& SP-HC   & 0.59 & 0.41 & 0.05 & 0.18 & 0.67 &  \bf 0.75 & -0.03 & 0.22 \\\cline{2-10}
& SP-M    & 0.33 & 0.66 &  \bf 0.68 & 0.65 & 0.40 & 0.22 & 0.63 & 0.61 \\\cline{2-10}
& SP-U    & 0.27 & 0.25 & 0.20 & 0.08 & 0.28 & 0.13 &  \bf 0.30 & 0.24 \\\cline{2-10}
\hline   
\multirow{1}{*}{{\bf{Mean SR}}}
& -       & 0.38 & 1.07 & 1.05 & 1.11 & 0.41 & 0.66 & 1.04 &  \bf 1.30 \\ \hline\hline

\multirow{15}{*}{\rotatebox[origin=c]{90}{\bf{Calmar Ratio}}} 
& NYSE-N  & 0.18 & 0.92 & 1.21 &  \bf 1.24 & 0.20 & 0.87 & 1.11 &  \bf 1.24 \\\cline{2-10}
& NYSE-O  & 0.29 & 11.40 & 9.30 & 11.12 & 0.39 & 1.99 & 8.80 &  \bf 13.60 \\\cline{2-10}
& MSCI    & -0.05 & 1.83 & 2.00 & 2.22 & -0.05 & 1.01 & 1.42 &  \bf 2.35 \\\cline{2-10}
& TSE     & 0.42 & 2.94 & 1.02 & 1.58 & 0.33 & 1.73 & 2.58 &  \bf 3.31 \\\cline{2-10}
& DOW30   &  \bf 0.30 & -0.09 & 0.09 & 0.11 & 0.28 & 0.23 & 0.07 & 0.19 \\\cline{2-10}
& SP-IT   & 0.33 & 0.27 & 0.18 & 0.13 & 0.35 & 0.39 & 0.37 &  \bf 0.56 \\\cline{2-10}
& SP-CS   & 0.39 & 0.26 & 0.25 & 0.24 &  \bf 0.44 & 0.09 & 0.19 & 0.27 \\\cline{2-10}
& SP-CSe  & 0.42 & 0.14 & 0.34 & 0.24 & 0.31 & 0.22 & 0.00 &  \bf 0.52 \\\cline{2-10}
& SP-F    & 0.15 & 0.05 & 0.32 & 0.25 & 0.16 &  \bf 0.38 & 0.20 & 0.20 \\\cline{2-10}
& SP-I    & 0.26 & 0.23 & 0.20 & 0.29 & 0.28 &  \bf 0.72 & 0.31 & 0.24 \\\cline{2-10}
& SP-RE   & 0.18 & 0.20 & 0.31 & 0.26 & 0.18 & 0.24 &  \bf 0.40 & 0.14 \\\cline{2-10}
& SP-E    & 0.15 & 0.00 & 0.15 & 0.15 & 0.18 & 0.06 & 0.16 &  \bf 0.25 \\\cline{2-10}
& SP-HC   & 0.37 & 0.32 & 0.08 & 0.17 & 0.41 &  \bf 0.57 & 0.04 & 0.22 \\\cline{2-10}
& SP-M    & 0.20 &  \bf 0.51 & 0.50 & 0.50 & 0.26 & 0.16 & 0.45 & 0.43 \\\cline{2-10}
& SP-U    & 0.21 & 0.24 & 0.20 & 0.12 & 0.22 & 0.14 &  \bf 0.28 & 0.21 \\\cline{2-10}
\hline   
\multirow{1}{*}{{\bf{Mean CR}}}
& -       & 0.25 & 1.28 & 1.08 & 1.24 & 0.26 & 0.59 & 1.09 &  \bf 1.58 \\ \hline\hline

\multirow{15}{*}{\rotatebox[origin=c]{90}{\bf{Maximum Drawdown (\%)}}} 
& NYSE-N  &  \bf 56.90 & 77.10 & 92.50 & 89.30 & 63.90 & 91.90 & 90.70 & 89.30 \\\cline{2-10}
& NYSE-O  & 41.20 &  \bf 32.90 & 46.50 & 40.40 & 36.90 & 58.40 & 39.30 & 38.40 \\\cline{2-10}
& MSCI    & 64.60 & 55.30 & 48.10 & 49.40 & 64.40 & 41.90 & 47.90 &  \bf 41.20 \\\cline{2-10}
& TSE     &  \bf 29.90 & 64.90 & 81.40 & 76.70 & 33.50 & 72.60 & 65.80 & 61.10 \\\cline{2-10}
& DOW30   &  \bf 47.10 & 88.10 & 82.50 & 80.30 & 47.30 & 63.70 & 78.80 & 53.80 \\\cline{2-10}
& SP-IT   &  \bf 55.10 & 67.80 & 71.30 & 77.00 & 55.40 & 72.40 & 68.10 & 70.20 \\\cline{2-10}
& SP-CS   & 31.70 & 49.90 & 64.60 & 64.40 &  \bf 29.30 & 77.00 & 51.90 & 39.30 \\\cline{2-10}
& SP-CSe  &  \bf 52.50 & 73.60 & 58.50 & 76.80 & 52.80 & 77.00 & 81.80 & 82.00 \\\cline{2-10}
& SP-F    &  \bf 67.30 & 84.30 & 91.90 & 91.60 & 69.00 & 84.50 & 74.60 & 74.90 \\\cline{2-10}
& SP-I    & 55.10 & 67.50 & 71.80 & 74.80 & 54.60 &  \bf 53.70 & 71.90 & 76.10 \\\cline{2-10}
& SP-RE   &  \bf 61.90 & 72.40 & 65.70 & 71.10 & 62.60 & 73.20 & 67.70 & 73.00 \\\cline{2-10}
& SP-E    & 54.20 & 83.60 & 85.40 & 82.50 &  \bf 51.60 & 78.00 & 83.30 & 81.20 \\\cline{2-10}
& SP-HC   & 43.80 & 60.10 & 76.00 & 72.90 &  \bf 41.70 & 70.60 & 77.80 & 56.00 \\\cline{2-10}
& SP-M    & 61.40 & 63.60 & 67.70 & 65.90 &  \bf 54.90 & 84.30 & 59.00 & 73.20 \\\cline{2-10}
& SP-U    & 44.60 & 51.00 & 55.70 & 56.40 &  \bf 43.80 & 62.60 & 49.80 & 55.50 \\\cline{2-10}
\hline   
\multirow{1}{*}{{\bf{Mean MDD (\%)}}}
& -       & 51.15 & 66.14 & 70.64 & 71.30 &  \bf 50.78 & 70.79 & 67.23 & 64.35 \\ \hline
	
\end{tabular}
\end{table*}

\begin{table*}[t] 
\caption{ Sharpe ratio ($SR$), Calmar ratio ($CR$), and annualized percentage yield ($APY$) for several  datasets for different OLPS methods, for a trading commission rate $c_r$ of $0.025\%$. The best values of each index are highlighted in bold type.}

\label{Tab_financial_measures2}
\centering
\color{black}
\begin{tabular}{|c|c|c|c|c|c|c|c|c|c|}
\hline
\multirow{2}{*}{$\boldsymbol{Metrics}\ (c_r=0.025\%)$} & \multirow{2}{*}{\bf{Dataset}}  & \multicolumn{5}{c|}{\bf{Representative OLPS methods}} & \multicolumn{3}{c|}{\bf{EGAB proposals}} \\
\cline{3-10}
& & \bf{UBAH} 
& \bf{PAMR} & \bf{OLMAR} & \bf{RMR} & \bf{EG} & $\boldsymbol{EG+}$ & $\boldsymbol{EGAB\mbox{-}N}$ & $\boldsymbol{EGAB\mbox{-}P}$ 
\\
\hline\hline

\multirow{15}{*}{\rotatebox[origin=c]{90}{\bf{Annualized Percentage Yield (\%)}}} 
& NYSE-N  & 10.20 & 62.10 &  \bf 104.10 & 102.80 & 12.90 & 79.20 & 93.40 & 101.30 \\\cline{2-10}
& NYSE-O  & 11.80 & 350.20 & 410.60 & 426.20 & 14.20 & 111.90 & 328.70 &  \bf 503.40 \\\cline{2-10}
& MSCI    & -3.40 & 90.30 & 88.20 &  \bf 100.50 & -3.10 & 41.00 & 63.40 & 89.30 \\\cline{2-10}
& TSE     & 12.50 & 177.20 & 76.50 & 112.60 & 11.20 & 92.70 & 158.80 &  \bf 187.10 \\\cline{2-10}
& DOW30   & 14.30 & -12.80 & 3.60 & 4.10 & 13.00 & 13.50 &  \bf 30.00 & 5.80 \\\cline{2-10}
& SP-IT   & 18.30 & 12.30 & 8.60 & 5.60 & 19.30 & 26.60 & 23.90 &  \bf 39.10 \\\cline{2-10}
& SP-CS   & 12.50 & 7.20 & 11.90 & 10.80 &  \bf 12.80 & 6.20 & 10.10 & 10.60 \\\cline{2-10}
& SP-CSe  & 22.00 & 4.60 & 15.70 & 14.20 & 16.10 & 10.90 & 0.00 &  \bf 42.90 \\\cline{2-10}
& SP-F    & 10.00 & -0.90 & 24.50 & 17.50 & 10.90 &  \bf 29.50 & 14.90 & 14.70 \\\cline{2-10}
& SP-I    & 14.50 & 9.40 & 10.00 & 16.60 & 15.40 &  \bf 38.40 & 16.70 & 13.80 \\\cline{2-10}
& SP-RE   & 11.30 & 8.30 & 15.60 & 13.60 & 11.60 & 16.70 & 23.50 &  \bf 23.70 \\\cline{2-10}
& SP-E    & 7.90 & -4.80 & 8.70 & 7.70 & 9.10 & 3.50 & 12.00 &  \bf 19.20 \\\cline{2-10}
& SP-HC   & 16.10 & 13.20 & 2.60 & 8.00 & 16.90 &  \bf 39.20 & -1.00 & 10.80 \\\cline{2-10}
& SP-M    & 12.30 & 25.50 &  \bf 29.20 & 28.00 & 14.10 & 12.10 & 20.40 & 24.00 \\\cline{2-10}
& SP-U    & 9.40 & 6.10 & 6.80 & 2.50 & 9.60 & 7.70 & 13.20 &  \bf 13.70 \\\cline{2-10}
\hline   
\multirow{1}{*}{{\bf{Mean APY (\%)}}}
& -       & 11.98 & 49.86 & 54.44 & 58.05 & 12.27 & 35.27 & 53.87 &  \bf 73.29 \\ \hline\hline

\multirow{15}{*}{\rotatebox[origin=c]{90}{\bf Sharpe Ratio}} 
& NYSE-N  & 0.35 & 1.17 & 1.72 & 1.71 & 0.48 & 1.34 &  \bf 1.73 & 1.68 \\\cline{2-10}
& NYSE-O  & 0.50 & 6.83 & 7.31 & 7.52 & 0.73 & 2.08 & 6.57 &  \bf 8.93 \\\cline{2-10}
& MSCI    & -0.29 & 2.26 & 2.05 &  \bf 2.37 & -0.27 & 0.88 & 1.66 & 2.11 \\\cline{2-10}
& TSE     & 0.65 &  \bf 2.34 & 0.75 & 1.14 & 0.55 & 1.24 & 1.79 & 1.86 \\\cline{2-10}
& DOW30   & 0.53 & -0.47 & -0.01 & 0.00 & 0.45 & 0.25 &  \bf 0.82 & 0.05 \\\cline{2-10}
& SP-IT   & 0.59 & 0.20 & 0.10 & 0.04 & 0.63 & 0.48 & 0.54 &  \bf 0.68 \\\cline{2-10}
& SP-CS   & 0.55 & 0.10 & 0.24 & 0.20 &  \bf 0.58 & 0.07 & 0.22 & 0.30 \\\cline{2-10}
& SP-CSe  & 0.62 & 0.01 & 0.28 & 0.24 & 0.54 & 0.16 & -0.10 &  \bf 0.76 \\\cline{2-10}
& SP-F    & 0.22 & -0.09 & 0.30 & 0.19 & 0.22 &  \bf 0.39 & 0.30 & 0.29 \\\cline{2-10}
& SP-I    & 0.45 & 0.14 & 0.14 & 0.29 & 0.48 &  \bf 0.80 & 0.36 & 0.25 \\\cline{2-10}
& SP-RE   & 0.25 & 0.10 & 0.24 & 0.20 & 0.24 & 0.27 &  \bf 0.43 & 0.41 \\\cline{2-10}
& SP-E    & 0.16 & -0.20 & 0.10 & 0.08 & 0.20 & -0.01 & 0.18 &  \bf 0.34 \\\cline{2-10}
& SP-HC   & 0.59 & 0.24 & -0.03 & 0.09 & 0.66 &  \bf 0.73 & -0.13 & 0.19 \\\cline{2-10}
& SP-M    & 0.33 & 0.50 &  \bf 0.57 & 0.54 & 0.40 & 0.18 & 0.46 & 0.45 \\\cline{2-10}
& SP-U    & 0.27 & 0.06 & 0.08 & -0.04 & 0.28 & 0.10 & 0.28 &  \bf 0.29 \\\cline{2-10}
\hline   
\multirow{1}{*}{{\bf{Mean SR}}}
& -       & 0.38 & 0.88 & 0.92 & 0.97 & 0.41 & 0.60 & 1.01 &  \bf 1.24 \\ \hline\hline

\multirow{15}{*}{\rotatebox[origin=c]{90}{\bf{Calmar Ratio}}} 
& NYSE-N  & 0.18 & 0.79 & 1.12 &  \bf 1.14 & 0.20 & 0.86 & 1.00 & 1.05 \\\cline{2-10}
& NYSE-O  & 0.29 & 10.57 & 8.75 & 10.44 & 0.39 & 1.91 & 8.12 &  \bf 12.94 \\\cline{2-10}
& MSCI    & -0.05 & 1.62 & 1.81 & 2.01 & -0.05 & 0.97 & 1.30 &  \bf 2.12 \\\cline{2-10}
& TSE     & 0.42 & 2.71 & 0.93 & 1.45 & 0.33 & 1.28 & 2.37 &  \bf 2.96 \\\cline{2-10}
& DOW30   & 0.30 & -0.14 & 0.04 & 0.05 & 0.27 & 0.21 &  \bf 0.49 & 0.10 \\\cline{2-10}
& SP-IT   & 0.33 & 0.18 & 0.12 & 0.07 & 0.35 & 0.37 & 0.35 &  \bf 0.56 \\\cline{2-10}
& SP-CS   & 0.39 & 0.14 & 0.18 & 0.17 &  \bf 0.44 & 0.08 & 0.19 & 0.27 \\\cline{2-10}
& SP-CSe  & 0.42 & 0.06 & 0.26 & 0.18 & 0.30 & 0.13 & 0.00 &  \bf 0.52 \\\cline{2-10}
& SP-F    & 0.15 & -0.01 & 0.27 & 0.19 & 0.16 &  \bf 0.35 & 0.20 & 0.20 \\\cline{2-10}
& SP-I    & 0.26 & 0.14 & 0.13 & 0.22 & 0.28 &  \bf 0.70 & 0.23 & 0.20 \\\cline{2-10}
& SP-RE   & 0.18 & 0.11 & 0.22 & 0.17 & 0.18 & 0.23 &  \bf 0.35 & 0.33 \\\cline{2-10}
& SP-E    & 0.15 & -0.05 & 0.10 & 0.09 & 0.18 & 0.04 & 0.14 &  \bf 0.24 \\\cline{2-10}
& SP-HC   & 0.37 & 0.21 & 0.03 & 0.11 & 0.41 &  \bf 0.55 & -0.01 & 0.19 \\\cline{2-10}
& SP-M    & 0.20 & 0.35 &  \bf 0.42 & 0.41 & 0.26 & 0.14 & 0.34 & 0.32 \\\cline{2-10}
& SP-U    & 0.21 & 0.11 & 0.12 & 0.04 & 0.22 & 0.12 & 0.28 &  \bf 0.29 \\\cline{2-10}
\hline   
\multirow{1}{*}{{\bf{Mean CR}}}
& -       & 0.25 & 1.12 & 0.97 & 1.12 & 0.26 & 0.53 & 1.02 &  \bf 1.49 \\ \hline\hline

\multirow{15}{*}{\rotatebox[origin=c]{90}{\bf{Maximum Drawdown (\%)}}} 
& NYSE-N  &  \bf 56.90 & 78.60 & 93.20 & 90.40 & 63.90 & 91.90 & 93.30 & 96.40 \\\cline{2-10}
& NYSE-O  & 41.20 &  \bf 33.10 & 46.90 & 40.80 & 36.90 & 58.50 & 40.50 & 38.90 \\\cline{2-10}
& MSCI    & 64.60 & 55.80 & 48.70 & 49.90 & 64.40 & 42.20 & 48.60 &  \bf 42.00 \\\cline{2-10}
& TSE     &  \bf 29.90 & 65.50 & 82.20 & 77.70 & 33.50 & 72.30 & 67.00 & 63.30 \\\cline{2-10}
& DOW30   &  \bf 47.10 & 94.00 & 83.40 & 81.40 & 47.30 & 64.00 & 60.90 & 58.20 \\\cline{2-10}
& SP-IT   &  \bf 55.10 & 69.50 & 72.30 & 77.90 & 55.40 & 73.00 & 68.40 & 70.20 \\\cline{2-10}
& SP-CS   & 31.70 & 51.50 & 65.20 & 65.40 &  \bf 29.30 & 77.90 & 51.90 & 39.30 \\\cline{2-10}
& SP-CSe  &  \bf 52.50 & 74.30 & 59.60 & 77.70 & 52.80 & 82.10 & 81.80 & 82.00 \\\cline{2-10}
& SP-F    &  \bf 67.30 & 84.30 & 92.00 & 91.90 & 69.00 & 85.00 & 74.60 & 74.90 \\\cline{2-10}
& SP-I    & 55.10 & 68.10 & 74.70 & 75.20 &  \bf 54.60 & 55.20 & 72.90 & 70.40 \\\cline{2-10}
& SP-RE   &  \bf 61.90 & 72.60 & 71.90 & 78.10 & 62.60 & 73.20 & 67.40 & 70.70 \\\cline{2-10}
& SP-E    & 54.20 & 90.60 & 87.00 & 84.40 &  \bf 51.70 & 79.00 & 86.10 & 81.00 \\\cline{2-10}
& SP-HC   & 43.80 & 63.80 & 77.00 & 73.40 &  \bf 41.70 & 70.70 & 79.30 & 56.50 \\\cline{2-10}
& SP-M    & 61.40 & 72.00 & 69.30 & 68.00 &  \bf 54.90 & 85.40 & 60.00 & 74.20 \\\cline{2-10}
& SP-U    & 44.60 & 53.70 & 57.90 & 58.70 &  \bf 43.80 & 63.90 & 46.80 & 47.00 \\\cline{2-10}
\hline   
\multirow{1}{*}{{\bf{Mean MDD (\%)}}}
& -       & 51.15 & 68.49 & 72.09 & 72.73 &  \bf 50.79 & 71.62 & 66.63 & 64.33 \\ \hline

\end{tabular}
\end{table*}

\begin{table*}[t] 
\caption{ Sharpe ratio ($SR$), Calmar ratio ($CR$), and annualized percentage yield ($APY$) for several  datasets for different OLPS methods, for a trading commission rate $c_r$ of $0.1\%$. The best values of each index are highlighted in bold type.}

\label{Tab_financial_measures3}
\centering
\color{black}
\begin{tabular}{|c|c|c|c|c|c|c|c|c|c|}
\hline
\multirow{2}{*}{$\boldsymbol{Metrics}\ (c_r=0.1\%)$} & \multirow{2}{*}{\bf{Dataset}}  & \multicolumn{5}{c|}{\bf{Representative OLPS methods}} & \multicolumn{3}{c|}{\bf{EGAB proposals}} \\
\cline{3-10}
& & \bf{UBAH} 
& \bf{PAMR} & \bf{OLMAR} & \bf{RMR} & \bf{EG} & $\boldsymbol{EG+}$ & $\boldsymbol{EGAB\mbox{-}N}$ & $\boldsymbol{EGAB\mbox{-}P}$ 
\\
\hline\hline

\multirow{15}{*}{\rotatebox[origin=c]{90}{\bf{Annualized Percentage Yield (\%)}}} 
& NYSE-N  & 10.20 & 38.40 & 82.10 & 80.00 & 12.80 & 0.70 & 81.20 &  \bf 86.50 \\\cline{2-10}
& NYSE-O  & 11.80 & 283.70 & 351.00 & 362.50 & 14.10 & 102.50 & 281.90 &  \bf 450.50 \\\cline{2-10}
& MSCI    & -3.40 & 60.90 & 65.90 &  \bf 75.60 & -3.20 & -19.30 & 56.80 & 70.60 \\\cline{2-10}
& TSE     & 12.50 & 139.90 & 57.70 & 88.70 & 11.00 & 56.50 & 127.80 &  \bf 143.70 \\\cline{2-10}
& DOW30   &  \bf 14.30 & -25.90 & -7.70 & -7.90 & 12.90 & 10.00 & 7.00 & 7.00 \\\cline{2-10}
& SP-IT   & 18.30 & -4.00 & -2.70 & -6.00 & 19.20 & 22.80 & 19.50 &  \bf 34.60 \\\cline{2-10}
& SP-CS   & 12.50 & -9.00 & -0.10 & -1.60 & 12.80 & 4.80 & 10.10 &  \bf 18.90 \\\cline{2-10}
& SP-CSe  & 22.00 & -10.50 & 4.30 & 1.60 & 16.00 & 15.40 & 0.00 &  \bf 42.90 \\\cline{2-10}
& SP-F    & 10.00 & -15.70 & 10.80 & 3.90 & 10.80 &  \bf 24.60 & 14.90 & 14.90 \\\cline{2-10}
& SP-I    & 14.50 & -6.90 & -2.00 & 3.20 & 15.30 & 7.60 &  \bf 15.70 & 11.90 \\\cline{2-10}
& SP-RE   & 11.30 & -8.00 & 2.60 & 0.40 & 11.50 & 14.90 & 15.70 &  \bf 18.10 \\\cline{2-10}
& SP-E    & 7.90 & -18.80 & -3.30 & -4.80 & 9.00 & 0.20 & 5.30 &  \bf 13.60 \\\cline{2-10}
& SP-HC   & 16.10 & -3.40 & -8.20 & -4.10 & 16.80 &  \bf 36.50 & -11.10 & 4.50 \\\cline{2-10}
& SP-M    & 12.30 & 6.70 & 15.40 & 13.70 & 14.00 & 11.70 & 5.10 &  \bf 17.60 \\\cline{2-10}
& SP-U    & 9.40 & -10.10 & -4.80 & -9.10 & 9.60 & 5.90 &  \bf 12.70 & 7.20 \\\cline{2-10}
\hline   
\multirow{1}{*}{{\bf{Mean APY (\%)}}}
& -       & 11.98 & 27.82 & 37.40 & 39.74 & 12.17 & 19.65 & 42.84 &  \bf 62.83 \\ \hline\hline

\multirow{15}{*}{\rotatebox[origin=c]{90}{\bf Sharpe Ratio}} 
& NYSE-N  & 0.35 & 0.69 & 1.34 & 1.32 & 0.47 & -0.04 &  \bf 1.50 & 1.42 \\\cline{2-10}
& NYSE-O  & 0.50 & 5.52 & 6.24 & 6.39 & 0.72 & 1.92 & 5.59 &  \bf 7.97 \\\cline{2-10}
& MSCI    & -0.29 & 1.49 & 1.51 &  \bf 1.76 & -0.28 & -0.54 & 1.46 & 1.65 \\\cline{2-10}
& TSE     & 0.65 &  \bf 1.84 & 0.56 & 0.89 & 0.54 & 0.74 & 1.43 & 1.42 \\\cline{2-10}
& DOW30   &  \bf 0.53 & -0.84 & -0.30 & -0.31 & 0.45 & 0.16 & 0.12 & 0.12 \\\cline{2-10}
& SP-IT   & 0.59 & -0.19 & -0.15 & -0.22 &  \bf 0.62 & 0.39 & 0.49 & 0.59 \\\cline{2-10}
& SP-CS   & 0.55 & -0.42 & -0.12 & -0.17 &  \bf 0.57 & 0.03 & 0.22 & 0.48 \\\cline{2-10}
& SP-CSe  & 0.62 & -0.35 & 0.01 & -0.06 & 0.54 & 0.26 & -0.10 &  \bf 0.76 \\\cline{2-10}
& SP-F    & 0.22 & -0.35 & 0.10 & 0.00 & 0.21 &  \bf 0.31 & 0.30 & 0.30 \\\cline{2-10}
& SP-I    & 0.45 & -0.29 & -0.14 & -0.02 &  \bf 0.48 & 0.08 & 0.35 & 0.20 \\\cline{2-10}
& SP-RE   & 0.25 & -0.27 & -0.03 & -0.07 & 0.24 & 0.23 & 0.26 &  \bf 0.30 \\\cline{2-10}
& SP-E    & 0.16 & -0.52 & -0.15 & -0.18 & 0.19 & -0.08 & 0.03 &  \bf 0.21 \\\cline{2-10}
& SP-HC   & 0.59 & -0.20 & -0.27 & -0.18 & 0.66 &  \bf 0.67 & -0.40 & 0.01 \\\cline{2-10}
& SP-M    & 0.33 & 0.06 & 0.26 & 0.22 &  \bf 0.39 & 0.17 & 0.03 & 0.37 \\\cline{2-10}
& SP-U    & 0.27 & -0.43 & -0.25 & -0.38 & 0.28 & 0.05 &  \bf 0.36 & 0.10 \\\cline{2-10}
\hline   
\multirow{1}{*}{{\bf{Mean SR}}}
& -       & 0.38 & 0.38 & 0.57 & 0.60 & 0.41 & 0.29 & 0.78 &  \bf 1.06 \\ \hline\hline

\multirow{15}{*}{\rotatebox[origin=c]{90}{\bf{Calmar Ratio}}} 
& NYSE-N  & 0.18 & 0.46 & 0.87 & 0.86 & 0.20 & 0.01 & 0.86 &  \bf 0.89 \\\cline{2-10}
& NYSE-O  & 0.29 & 7.13 & 7.29 & 8.58 & 0.38 & 1.75 & 6.53 &  \bf 11.28 \\\cline{2-10}
& MSCI    & -0.05 & 1.06 & 1.30 & 1.47 & -0.05 & -0.25 & 1.19 &  \bf 1.59 \\\cline{2-10}
& TSE     & 0.42 &  \bf 2.08 & 0.68 & 1.10 & 0.33 & 0.74 & 1.82 & 2.02 \\\cline{2-10}
& DOW30   &  \bf 0.30 & -0.26 & -0.08 & -0.09 & 0.27 & 0.15 & 0.11 & 0.11 \\\cline{2-10}
& SP-IT   & 0.33 & -0.05 & -0.03 & -0.07 & 0.35 & 0.31 & 0.31 &  \bf 0.47 \\\cline{2-10}
& SP-CS   & 0.39 & -0.11 & 0.00 & -0.02 &  \bf 0.43 & 0.06 & 0.19 & 0.32 \\\cline{2-10}
& SP-CSe  & 0.42 & -0.12 & 0.06 & 0.02 & 0.30 & 0.23 & 0.00 &  \bf 0.52 \\\cline{2-10}
& SP-F    & 0.15 & -0.16 & 0.12 & 0.04 & 0.16 &  \bf 0.28 & 0.20 & 0.20 \\\cline{2-10}
& SP-I    & 0.26 & -0.08 & -0.02 & 0.04 &  \bf 0.28 & 0.11 & 0.27 & 0.16 \\\cline{2-10}
& SP-RE   & 0.18 & -0.09 & 0.03 & 0.00 & 0.18 & 0.20 & 0.22 &  \bf 0.25 \\\cline{2-10}
& SP-E    & 0.15 & -0.19 & -0.04 & -0.05 &  \bf 0.17 & 0.00 & 0.06 &  \bf 0.17 \\\cline{2-10}
& SP-HC   & 0.37 & -0.04 & -0.09 & -0.05 & 0.40 &  \bf 0.51 & -0.12 & 0.07 \\\cline{2-10}
& SP-M    & 0.20 & 0.08 & 0.19 & 0.18 & 0.25 & 0.13 & 0.07 &  \bf 0.28 \\\cline{2-10}
& SP-U    & 0.21 & -0.12 & -0.06 & -0.11 &  \bf 0.22 & 0.09 & 0.21 & 0.14 \\\cline{2-10}
\hline   
\multirow{1}{*}{{\bf{Mean CR}}}
& -       & 0.25 & 0.64 & 0.68 & 0.79 & 0.26 & 0.29 & 0.79 &  \bf 1.23 \\ \hline\hline

\multirow{15}{*}{\rotatebox[origin=c]{90}{\bf{Maximum Drawdown (\%)}}} 
& NYSE-N  &  \bf 56.90 & 83.50 & 94.80 & 93.50 & 64.00 & 96.50 & 94.30 & 97.20 \\\cline{2-10}
& NYSE-O  & 41.20 & 39.80 & 48.10 & 42.20 &  \bf 37.00 & 58.50 & 43.20 & 39.90 \\\cline{2-10}
& MSCI    & 64.60 & 57.40 & 50.60 & 51.50 & 64.40 & 77.30 & 47.80 &  \bf 44.40 \\\cline{2-10}
& TSE     &  \bf 29.90 & 67.20 & 84.30 & 80.50 & 33.60 & 76.10 & 70.10 & 71.20 \\\cline{2-10}
& DOW30   &  \bf 47.10 & 99.20 & 91.20 & 91.70 & 47.40 & 67.10 & 63.30 & 63.30 \\\cline{2-10}
& SP-IT   &  \bf 55.10 & 77.30 & 85.10 & 91.50 & 55.50 & 74.40 & 63.10 & 73.00 \\\cline{2-10}
& SP-CS   & 31.70 & 82.60 & 68.40 & 68.20 &  \bf 29.40 & 80.20 & 51.90 & 58.40 \\\cline{2-10}
& SP-CSe  &  \bf 52.50 & 88.20 & 68.40 & 80.30 & 52.80 & 67.60 & 81.80 & 82.00 \\\cline{2-10}
& SP-F    &  \bf 67.30 & 96.30 & 92.80 & 92.70 & 69.10 & 87.00 & 74.30 & 74.60 \\\cline{2-10}
& SP-I    & 55.10 & 89.40 & 88.40 & 77.90 &  \bf 54.70 & 66.80 & 58.70 & 72.20 \\\cline{2-10}
& SP-RE   &  \bf 61.90 & 92.90 & 87.40 & 90.50 & 62.70 & 74.20 & 71.40 & 71.60 \\\cline{2-10}
& SP-E    & 54.20 & 98.20 & 90.60 & 89.00 &  \bf 51.70 & 83.40 & 88.20 & 81.00 \\\cline{2-10}
& SP-HC   & 43.80 & 87.60 & 91.30 & 83.50 &  \bf 41.70 & 71.30 & 89.90 & 60.30 \\\cline{2-10}
& SP-M    & 61.40 & 88.60 & 80.70 & 76.70 &  \bf 54.90 & 88.90 & 75.90 & 63.80 \\\cline{2-10}
& SP-U    & 44.60 & 85.80 & 76.20 & 86.50 &  \bf 43.80 & 67.00 & 62.10 & 51.10 \\\cline{2-10}
\hline   
\multirow{1}{*}{{\bf{Mean MDD (\%)}}}
& -       & 51.15 & 82.27 & 79.89 & 79.75 &  \bf 50.85 & 75.75 & 69.07 & 66.93 \\ \hline

\end{tabular}
\end{table*}

\section{Conclusions}
We have proposed EGAB, a novel family of exponentiated gradient updates that represent a natural generalization of the widely-known EG algorithms. EGAB updates are derived from the optimization of loss functions regularized by a parameterized Alpha-Beta divergence. Their $\beta$ hyperparameter controls the deformation of the exponentiated gradient, while $\gamma$ determines the influence of the solution on the learning rate. Initially devised for unconstrained optimization problems, the iterations were later extended to normalized updates that guarantee the unitary $\ell_1$-norm of the solution. Our proposals have been shown to unify and extend the update directions of various existing methods, including several online portfolio selection algorithms. 
Simulation results have confirmed the usefulness of the generalized exponentiated gradient updates in OLPS problems, both for developing innovative momentum and mean-reversion strategies, and for accounting for the presence of proportional transaction costs. 
As a future extension, our long term objective is to explore the possibility to replace additive stochastic gradient descent algorithms by generalized exponentiated gradient updates in large-scale problems and applications.

\color{black}

\appendices
\section{Derivation of the generalized $EGAB\mbox{-}U$ update for the Alpha-Beta Divergence regularizer}\label{App-EG-AB} 
Consider the function to minimize 
\eq{
	\widehat J_t(\bw) = \widehat{F}(\bw^\alpha)
	+\frac{1}{\eta} D^{(\alpha,\beta)}_{AB}(\bw\|\bw_{t}) 
}
where $F(\bw^\alpha) = L(\bw)$ and
\eq{
	\widehat{F}(\bw^\alpha) = F(\bw_{t}^\alpha) +  \langle\ \nabla_{{\bi w}^\alpha} F(\bw_t^\alpha),\ (\bw^\alpha-\bw_t^\alpha)\ \rangle .
}

We would like to find the minima of the function with respect to $\bw$  by solving the equation
\eq{
	\nabla_{{\bi w}} \widehat J_t(\bw) = {\bf 0}. \label{Crit1ax}
}
For this purpose, we determine the partial derivatives of the right-hand-size of the equation
\eq{
	\frac{\partial\widehat J_t(\bw)}{\partial w_{i}}  
	&= \frac{1}{\eta}\frac{\partial D^{(\alpha,\beta)}_{AB}(\bw\|\bw_{t})}{\partial w_{i}}   
	+  \frac{\partial \widehat{F}(\bw^\alpha)}{\partial w_{i}},
	\ \ \ i=1,\ldots,N.
	\label{Jgradax}
}
Given AB-divergence definition, a straightforward computation shows that its partial derivative can be written as 
\eq{
	\frac{\partial D^{(\alpha,\beta)}_{AB}\left(\bw\|\bw_{t}\right)}{\partial w_{i}}
	=  w_{i,t}^{\alpha+\beta-1}\left(\frac{w_i}{w_{i,t}}\right)^{\alpha-1} \log_{1-\beta}\left(\frac{w_i}{w_{i,t}}\right).
	\label{Dgradax}
}
From the definition of $\widehat F(\bw^\alpha)$ in \eqref{Fdef}, we see that 
\eq{
	\frac{\partial \widehat{F}(\bw^\alpha)}{\partial w_i^\alpha} 
	= 
	\frac{\partial F(\bw_t^\alpha)}{\partial w_i^\alpha}  .
	\label{approx2ax}
} 
This property in combination with the chain rule, allows one to simplify the gradient of  \eqref{Fdef} in terms of the gradient of $L(\bw)$ at $\bw_t$, to obtain
\eq{
	\frac{\partial \widehat{F}(\bw^\alpha)}{\partial w_i} 
	&= 	\frac{\partial w_i^\alpha} {\partial w_i}
	\frac{\partial  \widehat{F}(\bw^\alpha)} {\partial w_i^\alpha}\\
	&= 	\alpha w_i^{\alpha-1} \left(\frac{1}{\alpha w_{i,t}^{\alpha-1}}  
	\frac{\partial F(\bw_{t}^\alpha)}{\partial w_i}\right)\\	
	&=  \left(\frac{w_i}{w_{i,t}}\right)^{\alpha-1}  
	\frac{\partial L(\bw_{t})}{\partial w_i}.
	\label{Fgradax}			
}
After the substitution of \eqref{Dgradax} and \eqref{Fgradax} in \eqref{Jgradax}, we equate it to zero, to obtain 
\eq{
	0= \frac{\partial \widehat J_t(\bw) }{\partial w_i} 
	&= \frac{\partial  D^{(\alpha,\beta)}_{AB}\left(\bw\|\bw_{t}\right)}{\partial w_{i}} + \eta 	\frac{\partial \widehat{F}(\bw)}{\partial w_i} \\
	&= 
	\frac{1}{\eta} w_{i,t}^{\alpha+\beta-1}\left(\frac{w_i}{w_{i,t}}\right)^{\alpha-1}
	\log_{1-\beta}\left(\frac{w_i}{w_{i,t}}\right) \nonumber\\
	&\hspace{0.9cm} 
	+\left(\frac{w_i}{w_{i,t}}\right)^{\alpha-1}  
	\frac{\partial L(\bw_{t})}{\partial w_i} 
}
The solution of this equation provides a critical point of $\widehat J_t(\bw)$, which is attained at
\eq{
	w_{i,\star} = w_{i,t} \exp_{1-\beta} \left(-\eta\, w_{i,t}^{1-(\alpha+\beta)}\, \frac{\partial L(\bw_{t})}{\partial w_i} \right)
}
or equivalently, in vector form, as
\eq{
	\bw_{\star} 
	&= \bw_{t} \odot \exp_{1-\beta} \left(-\bm{\eta}_t \odot \nabla_{\bi w}L(\bw)  \right)
	\label{sol1b2}
}
where $\bm{\eta}_t=[\eta_{1,t},\ldots,\eta_{N,t}]^T$ and $\eta_{i,t} = \eta\, w_{i,t}^{1-(\alpha+\beta)} > 0$.

\section{Derivation of the $EGAB\mbox{-}N$ iteration}\label{App-EGn}

In this section we derive the normalized $EGAB\mbox{-}N(\alpha,\beta)$ updates. First for scale-invariant criteria, and later for the more general case of arbitrary differentiable loss functions.  

\subsubsection{The update for scale-invariant loss functions}

Consider a continuously differentiable, positive homogeneous loss function  
$L_I(\bw)$ which is scale-invariant, i.e., that satisfies the following property
\eq{
	L_I(c\, \bw)= c^0 L_I(\bw) = L_I(\bw)\quad \forall c\in \mathbf{R}_+.
}
By Euler's homogeneous function theorem (see \cite{Apostol}), a continuously differentiable and positive homogeneous function $L_I(\bw)$ of degree $k$ satisfies
\eq{
	\langle \nabla_{\bi w} L_I(\bw), \bw \rangle\ \dot =\ \bw^T \nabla_{\bi w} L_I(\bw) = k\, L_I(\bw) .
	\label{Euler}
}
\color{black}
Since a scale-invariant function is positive homogeneous with degree $k=0$, its gradient is orthogonal to the current estimate
\eq{
	\nabla_{\bi w} L_I(\bw) \perp \bw\,.
	\label{Perp1}
}
Because of the scale-invariance, the normalization of $\bw_\star$ does not modify the value of the loss function $L_I(\bw_\star)$, so each gradient descent update that yields a normalized value of $\bw_\star$ can be written as the following two step update 
\eq{
	\bw_{\star} &= \bw_{t} \odot \exp_{1-\beta} \left(-\bm{\eta}_t \odot \nabla_{\bi w}L_I(\bw_t) \right) \label{Iter1-1}\\
	\bw_{t+1}   &=  \bw_{\star}/\|\bw_{\star}\|_1 \, , \label{Iter1-2}
}
where the vector of learning rates is given by $\bm{\eta}_t=\eta\, \bw_t^\gamma$.

\subsubsection{The update for arbitrary differentiable functions} 

Consider now  any differentiable loss  function $L(\bw)$.
Through the normalization of its argument  to be of unit $\ell_1$ norm, we can always obtain the scale-invariant version of the loss function
\eq{
	L_I(\bw)=L\left(\frac{\bw}{\|\bw\|_1}\right),
	\label{LI}
}
which will be considered as a more convenient form to optimize. Let us use $\bI$ to denote the identity matrix, and also express the vector of unit elements as ${\bf 1}=(1,\ldots,1)^T$. Again, by Euler's homogeneous function theorem, and as a consequence of  equality \eqref{Euler}, it can be shown that the gradient of the loss 
\eq{
	\nabla_{\bi w} L_I(\bw_t)=   (\bI - \bm{1}\, \bw_t^T/\|\bw_t\|_1)/\|\bw_t\|_1  \nabla_{\bi w} L(\bw_t/\|\bw_t\|_1)
	\label{Grad2}
}
satisfies \eqref{Perp1}  and, consequently, is orthogonal to the current estimate $\bw_t$.
Because of this, the substitution of \eqref{Grad2} into \eqref{Iter1-1}-\eqref{Iter1-2}
will optimize the scale-invariant criterion in \eqref{LI}. However, since the normalization in \eqref{Iter1-2} at the iteration step $t-1$ already enforces $\|\bw_{t}\|_1=1$, the evaluation of the gradient in \eqref{Grad2} is simplified. After substituting it in \eqref{Iter1-1}, the minimization of the scale-invariant loss $L_I(\bw)$ is expressed in terms of $\nabla_{\bi w} L(\bw_t)$, the gradient of the initial differentiable loss, resulting the more general form of the  iteration
\eq{
	&\!\!\!\bw_{\star} \!
	=\! \bw_{t}\odot \exp_{1-\beta}\!\left(-\bm{\eta}_t\! \odot\! \nabla_{\bi w}L(\bw_t)+ \bm{\eta}_t  \bw_t^T \nabla_{\bi w}L(\bw_t) \right) \label{LI-1}\\
	&\!\!\!\bw_{t+1}   =  \bw_{\star}/\|\bw_{\star}\|_1 \,.  \label{LI-2}
}

\color{black}
When the function $L(\bw)$ is already scale-invariant, then it holds true that $\bw_t^T \nabla_{\bi w}L(\bw_t) = 0$. This cancels the second term of the gradient in \eqref{LI-1} and, as expected, \eqref{LI-1}-\eqref{LI-2} coincide with the updates \eqref{Iter1-1}-\eqref{Iter1-2} for scale-invariant functions.

\section{Alternative form of the $EGAB\mbox{-}N$ iteration}\label{App-2step}
Given the generalized EG iteration
\eq{
	\bw_{\star} &= \bw_{t} \odot \exp_{1-\beta} \left(-\bm{\eta}_t \odot \nabla_{\bi w}L_I(\bw_t) \right) \label{Iter1-1ap}\\
	\bw_{t+1}   &=  \bw_{\star}/\|\bw_{\star}\|_1 \, \label{Iter1-2ap}
}
we would like show its equivalence with the alternative expression for the update
\eq{
	\bw_{t+1} &=  \bw_{t} \odot \exp_{1-\beta}(-\bm{\eta}_t/\|\bw_{\star}\|_1^{\beta} \odot \nabla_{\bi w}L_I(\bw_t)\nonumber\\
	&\hspace{2.5cm} +\log_{1-\beta}\|\bw_{\star}\|_1^{-1}) \label{Iter5app}
}
where
\eq{
	\|\bw_{\star}\|_1 &= \bw_{t}^T \exp_{1-\beta} \left(-\bm{\eta}_t \odot \nabla_{\bi w}L_I(\bw_t) \right).
	\label{norma-1}
}
This is done by substituting \eqref{Iter1-1ap} into \eqref{Iter1-2ap} to finally obtain \eqref{Iter5app} as a direct consequence of the sequence of equivalences:  
\eq{
	&\hspace{-.4cm}\exp_{1-\beta}\left(-\bm{\eta}_t \odot \nabla_{\bi w}L_I(\bw_t) \right) /\|\bw_{\star}\|_1 \nonumber \\
	&\equiv  \left(\|\bw_{\star}\|_1^{-\beta}-\beta \|\bw_{\star}\|_1^{-\beta} \bm{\eta}_t \odot \nabla_{\bi w}L_I(\bw_t) \right)^{\frac{1}{\beta}} \nonumber \\
	&\equiv  \left(1-\beta \|\bw_{\star}\|_1^{-\beta} \bm{\eta}_t \odot \nabla_{\bi w}L_I(\bw_t)
	+\beta\frac{\|\bw_{\star}\|_1^{-\beta}-1}{\beta} \right)^{\frac{1}{\beta}} \nonumber \\	
	&\equiv  \exp_{1-\beta} \left(-\frac{\bm{\eta}_t}{\|\bw_{\star}\|_1^{\beta}} \odot \nabla_{\bi w}L_I(\bw_t) +\log_{1-\beta}\|\bw_{\star}\|_1^{-1}\right)\!\!.  \nonumber 
}

\section{Decomposition of the gradient into components that are parallel and orthogonal to the feasible manifold}\label{App_decomp}
One can observe in Figure \ref{Fig0-simplex} the unit $l_1$-norm constraint for nonnegative vectors. This manifold is the simplex defined by the intersection of the positive orthant with the oblique hyperplane, which is orthogonal to the $l_2$ normalized vector $\bar{\bf 1} \equiv \frac{1}{\sqrt{N}}{\bf 1}$ and passes through the uniform portfolio of unit $l_1$-norm, which we denote by $\bu=\tfrac{1}{N}{\bf 1}\equiv \frac{1}{\sqrt{N}} \bar{\bf 1}$. Hence, the orthogonal and parallel components of $\nabla_{\bi w}L_I(\bw_t) $ can be respectively obtained though the following set of projections
\eq{
	\nabla_{\bi w} L_{\perp}(\bw_t) 
	&= \bar{\bf 1}\bar{\bf 1}^T \nabla_{\bi w} L_I(\bw_t)\\
	\nabla_{\bi w} L_\parallel(\bw_t) 
	&= (\bI-\bar{\bf 1}\bar{\bf 1}^T) \nabla_{\bi w} L_I(\bw_t),
	\label{E2}
}
where, with  help of \eqref{eq_gradLI}, we can simplify \eqref{E2} as
\eq{
	\nabla_{\bi w} L_\parallel(\bw_t) 
	&=  \nabla_{\bi w} L_I(\bw_t)- (\bu^T \nabla_{\bi w} L_I(\bw_t)) {\bf 1}\\
	&=  \nabla_{\bi w} L(\bw_t)- \overline{\nabla_{\bi w} L(\bw_t)} {\bf 1},
}  
for 
\eq{
	\overline{\nabla_{\bi w} L(\bw_t)}  
	= {\bf u}^T \nabla_{\bi w}  L(\bw_t) 
	= \frac{1}{N}\sum_{i=1}^N \nabla_{w_i} L(\bw_t).
}



\section*{Authors contributions}
Authors A.C. and S.C. contributed equally to this article.
A.C. conceptualized the generalized Exponentiated Gradient framework, focusing on updates based on Alpha-Beta divergences and its subsequent application to Online Portfolio Selection problems. S.C. derived and formalized the generalized EG algorithms employing scale-invariant criteria and gradient projections. He also extended the algorithms to cover mean reversions approaches and accounted for transaction costs in the optimization process. A.S. played a crucial role in collecting data and performing experimental simulations to evaluate the performance of the algorithms. T.T. collaborated closely with A.C. on several versions of this manuscript and he investigated various potential applications of EGAB algorithms, including applications to OLPS problems and their limitations and perspectives.




\begin{IEEEbiography}[{\includegraphics[width=1in,height=1.25in,clip,keepaspectratio]{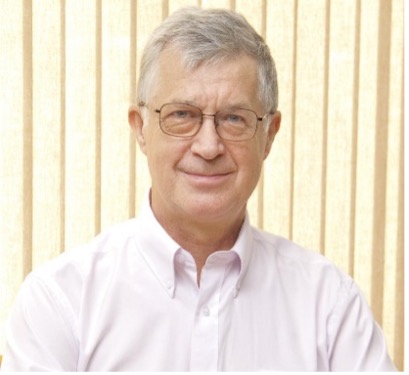}}]{Andrzej Cichocki} received the M.Sc. (with honors), Ph.D. and Dr.Sc. (Habilitation) degrees, all in electrical engineering from Warsaw University of Technology (Poland).
He was a Senior Team Leader and Head of the laboratories at RIKEN BSI (Brain Science Institute)  in Japan form 1995 till 2018. Currently he is a Professor in Systems Research Institute of Polish Academy of Science in Warsaw, and Nicolaus Copernicus University  in Torun, Poland. He is author of more than 500 technical journal papers and 6 monographs in English. He served as Associated Editor of three IEEE Transactions and he was founding Editor in Chief for Journal Computational Intelligence and Neuroscience. He is Fellow of the IEEE since 2013. Dr Cichocki work was honoured with several awards, including
2023, 2022, and 2021 Clarivate WoS Highly Cited Researcher.
\end{IEEEbiography}

\begin{IEEEbiography}[{\includegraphics[width=1in,height=1.25in,clip,keepaspectratio]{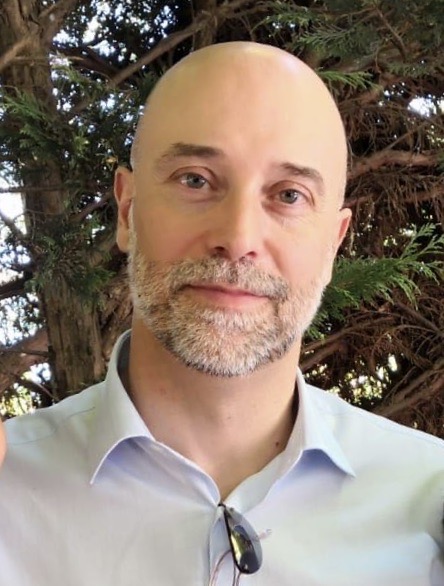}}]{Sergio Cruces}(S’93–M’99–SM’06) received the Graduate and Ph.D. degrees in telecommunications engineering from the University of Vigo (Spain), in 1994 and 1999, respectively. He is currently a Professor in the Department of Signal Theory and Communications, University of Seville (Spain), senior member of IEEE, and a member of the societies of SMCS, CIS, SPS, and ITS. His research interests include statistical signal processing, information theory, and machine learning. In 2015, he received the Best Paper award from the journal Entropy. Dr. Cruces has served as an Associate Editor of IEEE Transactions on Neural Networks and Learning Systems (2012-2015), of IEEE Transactions on Cybernetics (2015-2020), and as a member of the technical program committee of over forty international conferences.
\end{IEEEbiography}

\begin{IEEEbiography}[{\includegraphics[width=1in,height=1.25in,clip,keepaspectratio]{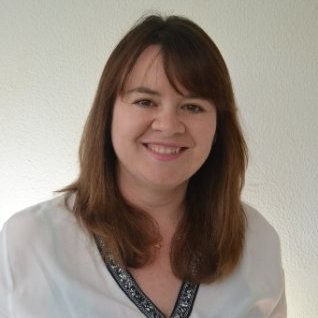}}]{Auxiliadora Sarmiento} received the M.Sc and Ph.D. degrees in telecommunications engineering from Universidad de Sevilla, Seville, Spain, in 2002 and 2011, respectively. Since 2005 she has been with the Department of Signal Theory and Communications, Universidad de Sevilla, where she  is currently an Associate Professor. She is senior member of IEEE and her current research interests include statistical signal processing, signal separation, information theory, bioengineering applications and EEG signal processing.
\end{IEEEbiography}

\begin{IEEEbiography}[{\includegraphics[width=1in,height=1.25in,clip,keepaspectratio]{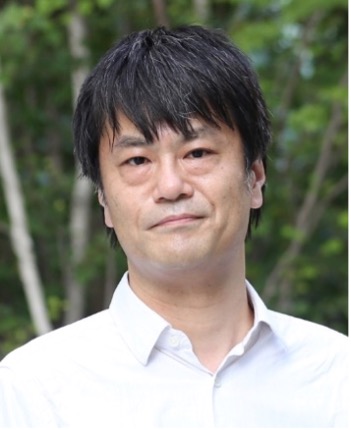}}]{Toshihisa Tanaka} received the B.E., the M.E., and the Ph.D. degrees from the Tokyo Institute of Technology in 1997, 2000, and 2002, respectively. In April 2004, he joined the Department of Electrical and Electronic Engineering, at the Tokyo University of Agriculture and Technology, where he is currently a Professor. 
He is a co-editor of two books, in 2008 and 2018. 
He served as editor-in-chief of Signals (MDPI) and as an associate editor and a guest editor of special issues in several journals, including IEEE Access, Neurocomputing, IEEE Transactions on Neural Networks and Learning Systems, and Neural Networks (Elsevier).
\end{IEEEbiography}
\vfill
\EOD
\end{document}